\documentclass{article}

 \usepackage[preprint]{neurips_2026}

\usepackage[utf8]{inputenc} 
\usepackage[T1]{fontenc}    
\usepackage{hyperref}       
\usepackage{url}            
\usepackage{booktabs}       
\usepackage{amsfonts}       
\usepackage{amsmath}        
\usepackage{nicefrac}       
\usepackage{microtype}      
\usepackage{xcolor}         
\usepackage{graphicx}
\usepackage{pifont}
\usepackage{multicol}
\usepackage{multirow}

\usepackage{wrapfig}
\usepackage{adjustbox}
\usepackage{colortbl}
\usepackage{enumitem}
\usepackage{fontawesome}
\usepackage{listings}
\usepackage{placeins}
\usepackage{titletoc}
\definecolor{lightgray}{gray}{0.95}
\newcommand{\cmark}{\textcolor{green}{\ding{51}}}%
\newcommand{\xmark}{\textcolor{red}{\ding{55}}}%

\title{RiskWebWorld: A Realistic Interactive Benchmark for GUI Agents in E-commerce Risk Management}

\author{%
  Renqi Chen \quad
  Zeyin Tao \quad
  Jianming Guo \quad Jing Wang\quad Zezhou Xu \\
  \textbf{Jingzhe Zhu}\quad \textbf{Qingqing Sun}\quad \textbf{Tianyi Zhang}\footnotemark[1] \quad \textbf{Shuai Chen}\thanks{Corresponding authors.} \\
  Ant International, Ant Group \\ 
  \texttt{\{chenrenqi.crq, zty113091, shuai.cs\}@ant-intl.com} \\
}

\begin{document}

\maketitle

\begin{abstract}
  Graphical User Interface (GUI) agents show strong capabilities for automating web tasks, but existing interactive benchmarks primarily target benign, predictable consumer environments. Their effectiveness in high-stakes, investigative domains such as authentic e-commerce risk management remains underexplored. To bridge this gap, we present RiskWebWorld, the first highly realistic interactive benchmark for evaluating GUI agents in e-commerce risk management. RiskWebWorld features 1,513 tasks sourced from production risk-control pipelines across 8 core domains, and captures the authentic challenges of risk operations on uncooperative websites, partially environmental hijackments. To support scalable evaluation and agentic reinforcement learning (RL), we further build a Gymnasium-compliant infrastructure that decouples policy planning from environment mechanics. Our evaluation across diverse models reveals a dramatic capability gap: top-tier generalist models achieve 49.1\% success, while specialized open-weights GUI models lag at near-total failure. This highlights that foundation model scale currently matters more than zero-shot interface grounding in long-horizon professional tasks. We also demonstrate the viability of our infrastructure through agentic RL, which improves open-source models by 16.2\%. These results position RiskWebWorld as a practical testbed for developing robust digital workers.
\end{abstract}

\section{Introduction}

Graphical user interface (GUI) agents have emerged as intelligent digital assistants capable of autonomously interacting with software environments~\cite{nguyen2025gui,gao2024assistgui,he2024webvoyager,wang2025ui,kumbhar2026towards}. By perceiving visual rendering states (e.g., screenshots) and underlying structural data (e.g., DOM trees), they translate natural language instructions into dynamic sequences of executable computer actions. As foundational models evolve, the application scope of GUI agents is expanding from normal consumer tasks~\cite{yao2022webshop,xie2024osworld,lee2024benchmarking} to high-stakes professional operations~\cite{peeters2025webmall,liao2025redteamcua,wang2026guiguard}. However, deploying agents in domains with strict operational constraints places severe demands on their reliability, long-horizon planning, and adaptability. To rigorously evaluate and advance their real-world capabilities, the community needs interactive benchmarks that reflect the multifaceted challenges of authentic commercial settings.

While current interactive benchmarks have established valuable evaluation protocols for general web browsing~\cite{zhou2023webarena,ye2026realwebassist,xue2025illusion,xu2025turkingbench}, mobile applications~\cite{rawles2405androidworld,kong2025mobileworld,li2026gui}, and desktop environments~\cite{wu2026marathon,yang2025macosworld,yang2025riosworld}, they exhibit critical limitations when scaling toward enterprise-grade applications. First, existing benchmarks predominantly simulate predictable, benign tasks, overlooking the complexities of dynamic risk analysis and cross-page verification. Second, specialized commercial domains, such as e-commerce risk management, remain underdeveloped from standardized evaluations. Crucially, most existing frameworks tightly couple policy planning and environment mechanics within closed loops. This architectural limitation restricts scalable benchmarking throughput and obstructs the fine-grained step orchestration required for stable agentic reinforcement learning (RL) training.

\begin{figure}
    \centering
    \includegraphics[width=0.99\textwidth]{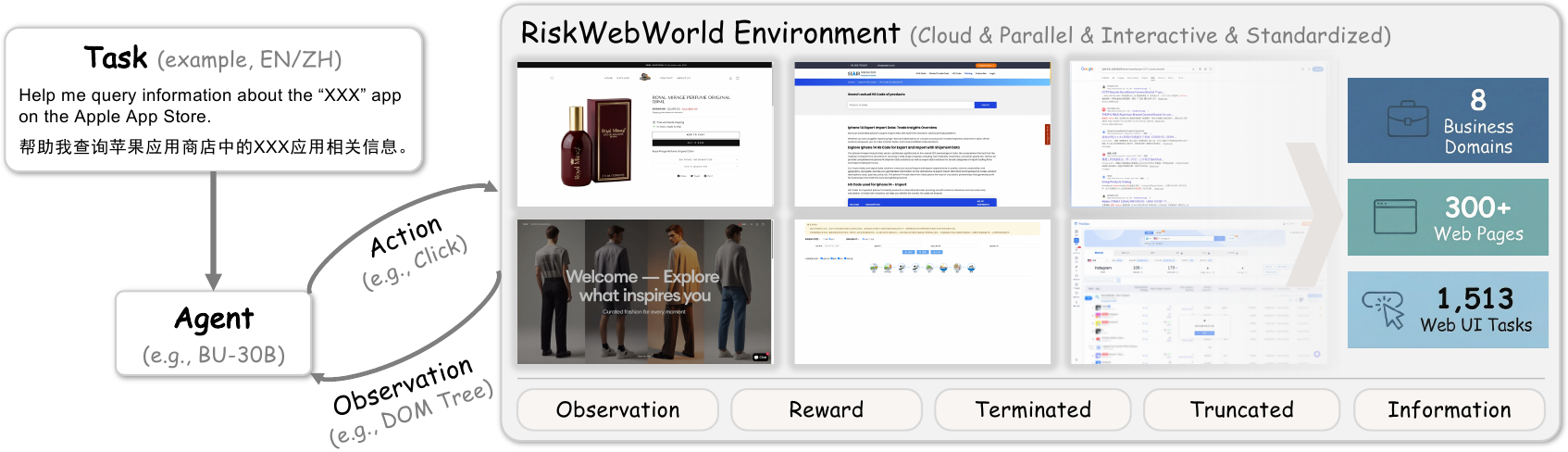}
    \vspace{-2mm}
    \caption{RiskWebWorld is a highly realistic, interactive benchmark for evaluating GUI agents in e-commerce risk management scenarios. It comprises 1,513 tasks across 8 business domains, with a scalable environment infrastructure that supports rigorous benchmarking and RL training.}
    \label{fig:introduction}
\end{figure}

Authentic e-commerce risk analysis~\cite{chen2025risk,zhang2025functionality} exemplifies these underexplored challenges. Agents operating in this domain cannot rely on straightforward, linear workflows. Instead, they act as proactive investigators, gathering heterogeneous data across multiple scattered and often uncooperative websites to uncover hidden risk signals. Throughout this process, agents frequently confront unpredictable environmental hijackments, including unexpected verification barriers (e.g., CAPTCHAs), disruptive pop-ups, and severe dynamic content shifts. Without testbeds that natively incorporate these sophisticated hurdles, it is difficult to gauge the functional readiness of GUI agents for live production deployment.
To bridge these gaps, we introduce RiskWebWorld (Figure \ref{fig:introduction}), the first highly realistic, interactive benchmark tailored for evaluating web agents in complex e-commerce risk management scenarios. Our main contributions are:
\begin{itemize}[leftmargin=*]
    \item We curate a comprehensive benchmark of 1,513 interactive Web UI tasks spanning 8 real-world business domains. Sourced directly from active online production pipelines, these tasks inherently include genuine environmental hijackments and unstructured verification challenges.
    \item We develop a scalable, Gymnasium-compliant environment infrastructure powered by CDP-based remote orchestration. This design decouples the agent's decision-making policy from the rendering mechanics, facilitating parallelized benchmarking and unlocking native support for agentic RL.
    \item We conduct an extensive evaluation of diverse baselines, including proprietary, open-weights, and GUI-specific foundation models, revealing clear performance tiers and systematic failure patterns that expose current limitations in open-ended exploration and multi-page evidence composition.
    \item We empirically demonstrate the viability of our infrastructure for advancing agent capabilities through RL, showing that open-source models (e.g., Qwen3-VL-8B) achieve consistent performance gains of up to 16.2\% when trained interactively in our environment.
\end{itemize}

Our evaluation reveals that \textbf{foundational model scale overwhelmingly dictates performance in complex web environments, dwarfing the impact of specialized zero-shot interface grounding}. Top-tier generalist models (e.g., Gemini-3-Pro, GPT-5.2) reach success rates near 50\%, whereas specialized GUI models face near-total failure (often 0\% success) due to action misrouting and argument hallucination. Even the most capable agents, however, face strict performance ceilings when required to execute continuous deductive verification sequences across scattered pages. Ultimately, our systematic evaluation and agentic RL demonstration establish RiskWebWorld as a critical testbed for transforming web agents from constrained assistants into robust digital workers.

\section{Related Works}
\paragraph{GUI Agents}
GUI agents are intelligent systems powered by large language or vision-language models (LLMs/VLMs) that perceive graphical user interfaces and execute automated actions to accomplish user-specified tasks. Early approaches primarily relied on \textit{expert workflows} with modular planner-actioner architectures~\cite{li2024appagent,zhao2025cola,wang2025mobile,xie2025scaling,zhang2025ufo2,jiang2025appagentx}, which often suffer from error accumulation in long-horizon tasks. Recently, research has broadly converged into two paradigms: (1) \textit{data-driven fine-tuning}, where multimodal LLMs are trained end-to-end on GUI datasets~\cite{lin2025showui,luo2025gui,zhou2025gui,gao2026ui,qin2025ui,tang2025gui,wang2025ui}, though they still face generalization bottlenecks on unseen scenarios; (2) \textit{versatile GUI agent frameworks}~\cite{browser_use2024,openmanus2025,openai_cua,claude_cua,xu2026mobile} that seamlessly integrate foundation models with tailored tools and enhanced context management to support robust real-world interactions and trajectory collection. These developments underscore the promise of GUI agents for open-ended web automation, naturally motivating the critical need for highly realistic benchmarks to evaluate their evolving capabilities.

\paragraph{Benchmarks for GUI Agents}
Benchmarks for evaluating GUI agents comprise two categories: static and interactive evaluators. Static (or offline) benchmarks rely on constructed datasets that provide paired GUI states (e.g., historical screenshots or DOM trees) alongside ground-truth next-action annotations~\cite{mialon2023gaia,deng2023mind2web,kapoor2024omniact,rawles2023androidinthewild,im2025modular,peeters2025webmall}. While convenient for metric assessment, they inherently oversimplify real-world interfaces and neglect the compounding dynamics of sequential execution. Conversely, interactive (or online) benchmarks situate agents within live or simulated environments, allowing them to freely interact with the GUI over multiple turns and measuring success based on ultimate task completion. Although interactive benchmarks have been well-established for general web browsing~\cite{yao2022webshop,pan2024webcanvas,xue2025illusion,gan2026guideweb}, mobile applications~\cite{gu2026generalization,liu2025learnact,rawles2405androidworld,li2026gui} and desktop environments~\cite{xie2024osworld,guo2025susbench,yang2025macosworld,sumyk2026cuaaudit,wu2026marathon}, they primarily evaluate agents in relatively benign environments focused on everyday consumer tasks (e.g., buying a product, fetching simple information). This focus leaves a critical gap in assessing agent capabilities for complex, high-stakes professional workflows. Moreover, while a scarce amount of preliminary research touches upon highly specialized commercial domains~\cite{chen2025risk,peeters2025webmall} like e-commerce risk management, systematic evaluation remains absent.
\vspace{-1mm}
\paragraph{Gaps and Contributions}
Current benchmarks broadly fail to reflect the stringent demands of highly specialized commercial production settings. Authentic risk analysis introduces severe domain-specific hurdles: agents must gather scattered heterogeneous data across multiple websites to uncover deeply hidden insights, while simultaneously navigating specialized out-of-distribution interfaces and unpredictable environmental constraints (e.g., CAPTCHAs, pop-ups). Because current benchmarks omit these challenges, they cannot adequately ensure the real-world applicability of UI agents. We therefore develop RiskWebWorld, a highly realistic, interactive e-commerce risk management benchmark coupled with a scalable environment infrastructure, to specifically bridge these gaps.

\vspace*{-1mm}
\section{RiskWebWorld Benchmark Infrastructure}\label{sec:RiskUI}
\begin{figure}
    \centering
    \includegraphics[width=0.99\textwidth]{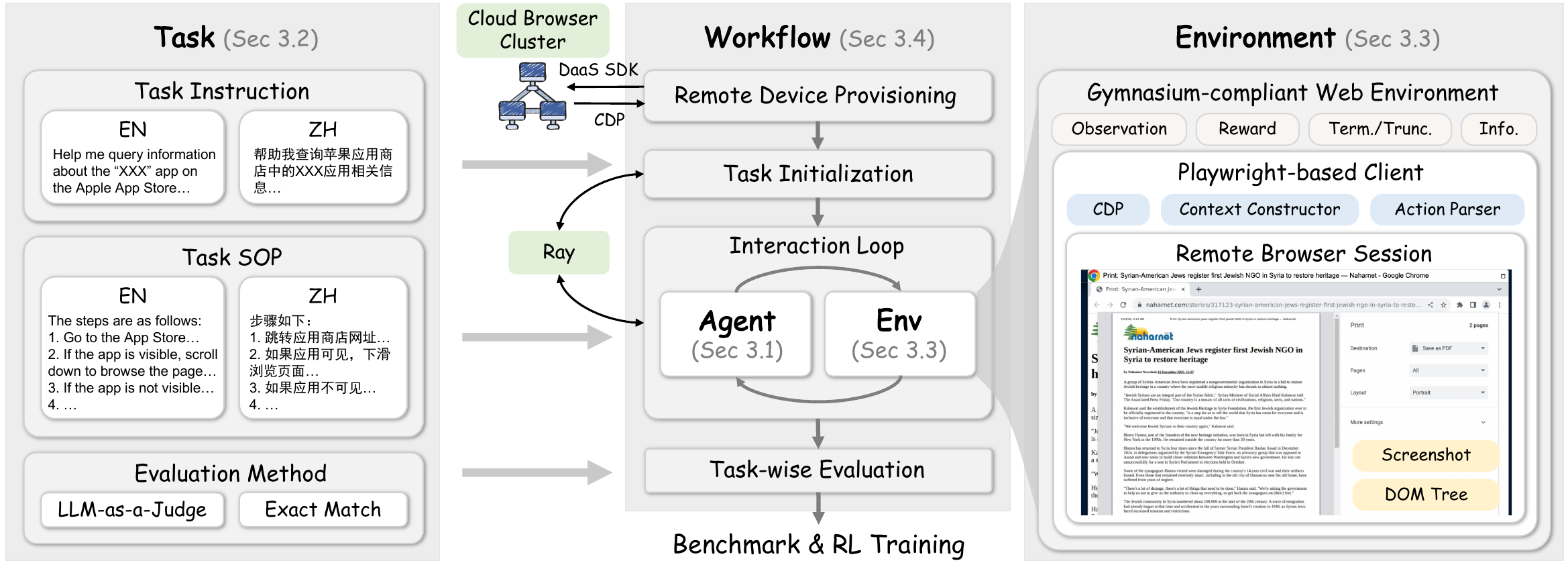}
    \vspace{-2mm}
    \caption{RiskWebWorld benchmark infrastructure. Benchmark procedure is defined by the workflow (Sec.~\ref{sec:workflow}): First, remote browser devices are provisioned in the cloud and connected to the environment module through CDP. Then the environment is initialized with the task configuration (Sec.~\ref{sec:task_configuration}), and the agent (Sec.~\ref{sec:agent}) starts to interact with the environment (Sec.~\ref{sec:environment}) in a step-wise manner. After the loop terminates or truncates, evaluation results could serve for both benchmarking and training.}
    \label{fig:architecture}
\end{figure}

\vspace*{-1mm}
Benchmarking the capability of a GUI agent in live website production settings requires handling the interaction between the agent and the website environment towards the target task. RiskWebWorld realizes this by (1) an agent module that orchestrates the multi-turn interaction and decouples the agent's decision making from the environment, (2) a suite of tasks that define the target tasks and the corresponding evaluation metrics, (3) an environment module that supports parallel, isolated, and stable website interactions, and (4) a specialized workflow that drives the evaluation process. The overall architecture of the RiskWebWorld benchmark infrastructure is illustrated in Figure~\ref{fig:architecture}.

\subsection{Agent}\label{sec:agent}
The agent's interaction with the GUI environment is typically formulated as a partially observable Markov decision process (POMDP)~\cite{xie2024osworld,yang2025macosworld} with an environment state space $S$, an observation function $O: S \rightarrow O$, an action space $A$, a transition function $T: S \times A \rightarrow S$, and a reward function $R: S \times A \rightarrow {R}$. In each turn, the agent receives an observation $o_t\in O$ (e.g., DOM tree, screenshot) from the website, and selects an action $a_t\in A$ (e.g., click, type) to execute. The environment then transitions to a new state $s_{t+1}=T(s_{t}, a_t)$, and the agent receives a reward $r_t=R(s_t, a_t)$. The loop continues until (1) terminated: the agent declares that the task is completed (success or failure), or (2) truncated: the interaction reaches a maximum number of steps. 

In RiskWebWorld, we decouple the agent's decision making from the environment by maintaining a five-tuple of the interaction state for the agent: $\left \langle o_t, r_{t-1}, Terminated, Truncated, Info \right \rangle$, where $Terminated$ and $Truncated$ are boolean flags indicating whether the interaction is terminated or truncated, and $Info$ is a dictionary that contains additional information (e.g., action validation). The role of the agent module is to monitor the interaction process carefully.

\subsection{Task Configuration}\label{sec:task_configuration}
Each task in RiskWebWorld is defined by a task configuration consisting of three components. (1) Task instruction: a natural language description of the task, comprising the task's goal and output format constraints. (2) Task standard operating procedure (SOP): a step-by-step procedure that guides the agent to complete the task. The SOP is annotated by human experts and serves as a reference for the agent to follow, which directly decides the task's difficulty level. (3) Task evaluation method: each task is associated with a specific evaluation method that quantifies the agent's performance on the task. For example, the evaluation method can be either LLM judgment-based or exact match-based, depending on the nature of the task. Figure~\ref{fig:task_example} illustrates a real example of a task configuration.

\vspace*{-1mm}
\subsection{Environment}\label{sec:environment}
Existing web-agent frameworks (e.g., Browser Use~\cite{browser_use2024}) are typically designed as closed-loop, autoregressive systems aimed at inference-time task execution for end users (consumer-facing). In such autonomous paradigms, the agent encapsulates both the policy planning and the environment execution within an opaque, uninterruptible monolithic loop. To unlock transparent, lightweight, and fine-grained evaluation for rigorous benchmarking, we systematically rebuild the web agent framework to formulate a scalable, Gymnasium-compliant environment, which also synergistically fulfills the strict, step-wise trajectory sampling requirements of RL training for GUI agents. Table~\ref{tab:comparision} summarizes the comparison between our RiskWebWorld environment and existing web-agent environments. Three key features of the environment module are as follows.

\vspace*{-1mm}
\paragraph{Unified MDP Decoupling}
We decoupled the environment mechanics from the autonomous policy planner based on two fundamental objectives:
(1) \textbf{Universal Benchmarking Standardization}: The core intent is to elevate the environment from a framework-specific closed agent into a reproducible, model-agnostic benchmarking testbed. By strictly refactoring the monolithic framework to expose formalized Gymnasium-compliant MDP primitives—while maintaining its original execution semantics—we logically isolate the environment's state-transition mechanism from internal prompt engineering and LLM inferences. This standardizes the pipeline of extracting multi-modal observations and validating JSON action schemas, endowing researchers with the capacity to seamlessly "plug-and-play" diverse foundation models.
(2) \textbf{Lightweight Distributed Architecture}: Building on the environment above, rigorous benchmarking and RL training demand the concurrent execution of open-domain tasks. Standard frameworks collapse at this scale due to localized UI threads, and verbose I/O-blocking system logs. To overcome this, we deployed the environment as a lightweight execution layer inside Ray-backed remote actors. Each actor connects to dynamically provisioned cloud sandboxes through the remote Chrome DevTools Protocol (CDP), packages multimodal histories for the policy network, and executes externally generated actions. To further stabilize large-scale parallel runs, we introduce a streamlined fault-interception mechanism that safely converts volatile Pydantic validation and DOM parsing exceptions into silent boolean failure signals, thereby preventing cascading worker crashes, reducing memory overhead, and increasing transition throughput.

\vspace*{-1mm}
\paragraph{CDP-Based Remote Orchestration}
At the browser-execution layer, we eliminate local browser instantiation entirely and instead interact with cloud-hosted Chromium instances through the CDP. At the beginning of each episode, the environment uses Daas~\cite{gao2026ui} SDK to provision an isolated browser session and obtain a low-level WebSocket endpoint associated with a unique Trace ID. A Playwright-based client then attaches directly to this endpoint, allowing high-level Python-side DOM control to operate transparently over the remote rendering engine. Furthermore, this CDP-driven initialization enforces an execution profile (e.g., a fixed 1920$\times$1200 viewport, and neutral proxy injections). These controls reduce rendering-side stochasticity and minimize DOM serialization latency across the remote procedure call (RPC) boundary, thereby producing stable high-fidelity visual and textual observations for policy network.

\begin{table}[t!]
   \centering
    \caption{The comparison between our RiskWebWorld environment and existing web-agent environments. Our environment is the only one that satisfies all the criteria for a standardized, scalable, and robust web-agent environment, which is essential for rigorous benchmarking and RL training.}
    \vspace{-2mm}
    \begin{adjustbox}{width=\textwidth}
    \begin{tabular}{lcccccc}
    \toprule
    \multirow{2}{*}{WebUI Env} & \multirow{2}{*}{\shortstack{Dynamic \\ Interaction}} & \multirow{2}{*}{\shortstack{General \\ Action Space}} & \multirow{2}{*}{\shortstack{Standardized \\ MDP}} & \multirow{2}{*}{\shortstack{ Parallel\\ Scalability}} & \multirow{2}{*}{\shortstack{Local-Free \\ Execution}} & \multirow{2}{*}{\shortstack{Framework-\\ Decoupled}} \\
    & \\
    \midrule
    Mind2Web~\cite{deng2023mind2web} & \xmark & \cmark & \xmark & \xmark &\cmark & \xmark \\

    Online-Mind2Web~\cite{xue2025illusion} & \cmark & \cmark & \xmark & \xmark &\cmark & \xmark \\

    WebShop~\cite{yao2022webshop} & \cmark & \xmark & \cmark & \xmark &\xmark & \xmark \\

    WebArena~\cite{zhou2023webarena} & \cmark & \cmark & \cmark & \xmark & \xmark & \xmark \\

    WebServ~\cite{lu2025webserv} & \cmark & \cmark & \xmark & \cmark &\xmark & \xmark \\

    WebGym~\cite{bai2026webgym} & \cmark & \cmark & \cmark & \cmark &\xmark & \xmark \\

     \textbf{RiskWebWorld Env} & \cmark & \cmark & \cmark & \cmark & \cmark & \cmark \\
    \bottomrule    
    \end{tabular}
    \end{adjustbox}
    \label{tab:comparision}
\end{table}

\paragraph{Granular MDP Control} Dismantling the native autonomous run-loop grants the external orchestrator surgical time-step control. This discrete granularity not only allows the external algorithm to explicitly inject intermediate actions and gracefully handle terminal states, but naturally transforms our benchmark into a pristine RL native habitat. Most crucially, this explicit step sequencing enables the procedural injection of tailored shaping rewards (e.g., evaluation method querying trajectory accuracy), which assigns detailed credit across the sampled open-web exploratory paths.

\subsection{Workflow}\label{sec:workflow}
Given a task configuration, our workflow takes 4 steps to benchmark it: (1) \textbf{Remote Device Provisioning} first acquires an isolated cloud browser sandbox via Daas SDK, retrieving a low-level CDP debugging endpoint tied to a unique Trace ID. (2) \textbf{Task Initialization} then connects to this remote device via Playwright to initialize the task-specific environment, where task configurations are recorded and the initial observation is obtained. Then (3) \textbf{Interaction Loop} begins multi-round interaction explicitly between the external policy agent and the environment. When the interaction terminates or truncates, the workflow enters (4) \textbf{Task-wise Evaluation}. Depending on the task type, this stage may invoke different evaluation protocols, ultimately producing granular scalar rewards.

\section{RiskWebWorld Tasks}
RiskWebWorld comprises 1,513 Web UI tasks organized into two subsets: (1) \textbf{RiskWebWorld-Standard}, which contains 1070 tasks, includes both tasks accompanied by SOPs and tasks whose scenarios are not compatible with SOP-based execution. This subset is designed to evaluate an agent's capability in regular production settings.
(2) \textbf{RiskWebWorld-Challenge}, which contains 443 tasks, consists of tasks for which no executable SOPs are available, thereby requiring stronger long-horizon reasoning and exploration capabilities from the agent.
The benchmark spans 8 business domains, further narrowing the gap between general benchmarks and e-commerce risk management scenarios.

\subsection{Task Curation}\label{sec:data_curation}
Our raw data is collected from an internal online production pipeline in which commercial tasks are executed by various information-seeking agents (e.g., Web UI agents~\footnote{We deployed the state-of-the-art Qwen3.5-397B on the Browser Use framework, with a revised action space and prompt.} and Deep Research agents). For GUI agents, all tasks are restricted to be completed within 20 steps to satisfy efficiency constraints in real-world production settings. After collecting more than 10,000 task instances, we apply a rigorous filtering process to construct the RiskWebWorld-Standard subset.
The first filtering criterion is result correctness. For tasks with existing labels, we compare the agent outputs against the labels. If they are consistent, the labels are retained and tasks are preserved; otherwise, we verify the label validity by cross-checking against the actual website state after task execution, ensuring label accuracy. For tasks without labels, human annotators manually assess the outputs, retaining correct answers or re-annotating incorrect ones as labels.

We then apply three fine-grained criteria to ensure benchmark quality:
(1) Task uniqueness. Duplicate tasks with identical instructions and SOPs are removed to promote diversity within the benchmark.
(2) Task time invariance. Tasks that are highly time-sensitive and likely to become obsolete within a short period (e.g., tasks related to flash sales) are excluded to preserve the benchmark's long-term relevance.
(3) Task relevance. Since the production pipeline utilizes various agents with different capabilities, the correct answer may not be derived from GUI agents. Such tasks are manually removed to ensure that all tasks are compatible with the evaluation agent.

In total, we obtain 1,070 tasks for the RiskWebWorld-Standard subset. Among them, 443 tasks are accompanied by executable SOPs, while the remaining 627 tasks do not include SOPs due to the dynamic nature of their scenarios or the absence of fixed exploration paths for completion.
To further enrich the benchmark with more challenging and generalizable tasks, RiskWebWorld-Challenge subset is constructed by removing the SOP guidance from these 443 SOP-accompanied tasks by our annotators. This setting requires the agent to complete the task without procedural references, thereby placing greater demands on its task-level process understanding and planning capabilities.

\subsection{Task Statistics}
\begin{figure}
    \centering
    \includegraphics[width=\textwidth]{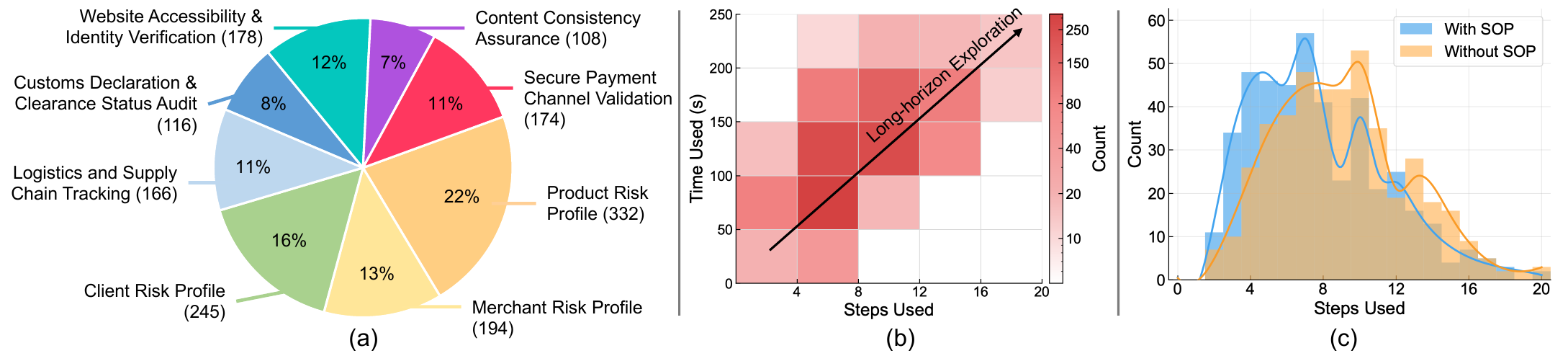}
    \caption{RiskWebWorld task statistics. (a) Task category distribution across 8 business domains. Exact values are given in parentheses. (b) Distributions of execution time and step count across task instances. (c) Comparison of step count distributions between SOP-accompanied and SOP-free tasks.}
    \label{fig:task_status}
\end{figure}

\paragraph{Category}
We adopted the task taxonomy proposed by previous work~\cite{chen2025risk} to encompass 8 business domains: Product Risk Profile (product-specific risk attributes; 332 tasks), Merchant Risk Profile (legal registration details, business licenses, etc.; 194 tasks), Client Risk Profile (publicly available identifiers, etc.; 245 tasks), Logistics and Supply Chain Tracking (shipping status through courier, freight, or e-commerce platforms; 166 tasks), Customs Declaration \& Clearance Status Audit (116 tasks), Website Accessibility \& Identity Verification (178 tasks), Content Consistency Assurance (108 tasks), and Secure Payment Channel Validation (174 tasks). The detailed task distribution across these categories is illustrated in Figure~\ref{fig:task_status} (a). The top three most heavily represented categories are Product Risk Profile, Client Risk Profile, and Merchant Risk Profile, highlighting a strong emphasis on core entity-level risk assessment.

\paragraph{Runtime Profiling}
For tasks collected from the online production pipeline, all execution metadata, including execution time and the number of interaction steps, are logged by our internal monitoring system, as illustrated in Figure~\ref{fig:frontend}. We re-perform the RiskWebWorld-Challenge tasks on the same Web UI agent settings as RiskWebWorld-Standard to obtain the corresponding execution metadata. Then we analyze the distributions of execution time and step count across all task instances in Figure~\ref{fig:task_status}(b). In general, tasks involving more interaction steps impose greater demands on long-horizon exploration and planning. We observe a positive correlation between execution time and step count, where the task distribution also exhibits a noticeable upper-left trend: tasks with the same number of steps need longer execution times. This pattern is attributed to factors such as website load latency and LLM inference time (longer interaction paths involve longer histories).

\paragraph{SOP Guidance}
To quantify the impact of SOP guidance on task difficulty, we compare the distributions of step counts between SOP-accompanied (RiskWebWorld-Standard) and SOP-free tasks (RiskWebWorld-Challenge), as shown in Figure~\ref{fig:task_status}(c). SOP-free tasks generally require more interaction steps, suggesting that the absence of procedural guidance increases task complexity.

\paragraph{Environment Hijackments}

RiskWebWorld tasks comprise a range of environmental hijackments that evaluate GUI agents' ability to handle real-world web interaction challenges. As summarized in Figure~\ref{fig:task_challenge}, we identify three representative types of hijackments: (1) Verification Barrier, where the agent needs to overcome human verification mechanisms (e.g., CAPTCHA) to access the target information; (2) Unpredictable Pop-up, where the agent must deal with unexpected pop-ups (e.g., cookie consent forms or region selection prompts) that disrupt the normal interaction flow; and (3) Dynamic Content Shift, where the agent should adapt to dynamic changes in website content (e.g., website load latency or redirection) that affect the availability or accessibility of information.

\newcommand{\ccp}{\cellcolor{blue!10}}
\definecolor{light_blue}{RGB}{209, 238, 255}
\newcommand{\ccb}{\cellcolor{light_blue}}

\begin{table*}[!t] 
    \centering
    \caption{Performance of baseline agents on the RiskWebWorld by task category. The columns delineate abbreviations: PRP (Product Risk Profile), MRP (Merchant Risk Profile), CRP (Client Risk Profile), LSCT (Logistics and Supply Chain Tracking), CDCSA (Customs Declaration \& Clearance Status Audit), WAIV (Website Accessibility \& Identity Verification), CCA (Content Consistency Assurance), SPCV (Secure Payment Channel Validation). The Overall column reports the success rate (SR) over 1,513 tasks. Highest SRs in each category are in \colorbox{blue!10}{purple}; second highest in \colorbox{light_blue}{blue}.}
    \begin{adjustbox}{width=\textwidth}
    \begin{tabular}{lccccccccc}
    \toprule
    \multirow{2.5}{*}{Model} & \multicolumn{8}{c}{Commercial Domain} & \multirow{2.5}{*}{Overall} \\ \cmidrule{2-9}
    & PRP & MRP & CRP & LSCT & CDCSA & WAIV & CCA & SPCV & \\
    \midrule
    \rowcolor{gray!15}
    \multicolumn{10}{l}{\textit{Commercial Models}} \\
    GPT-5.2 & \ccp 49.7 & \ccb 35.6 & \ccb 54.7 & 48.8 & 62.1 & 56.7 & \ccp 54.6 & 32.2 & \ccb 48.7 \\
    Gemini-3-Pro & 39.2 & \ccp 38.7 & 53.5 & \ccp 56.6 & \ccp 69.8 & \ccp 60.7 & 50.9 & \ccp 39.1 & \ccp 49.1 \\
    Claude-Sonnet-4.5 & 22.3 & 30.9 & 49.0 & 30.1 & 51.7  & 31.5 & 27.8 & 23.0 & 32.4 \\
    \multicolumn{1}{l}{\textit{Average}} & 37.1 & 35.1 & 52.4 & 45.2 & 61.2 & 49.6 & 44.4 & 31.4 & 43.4 \\
    \midrule
    \rowcolor{gray!15}
    \multicolumn{10}{l}{\textit{High-capability Open-source Models}} \\
    Kimi-K2.5 & 18.1 & 34.0 & \ccp 56.3 & \ccb 50.0 & 44.8 & 48.3 & 7.4 & 0.0 & 32.6 \\
    GLM-4.6V & 1.8  & 10.3 & 1.6 & 0.0 & 0.0 & 23.0 & 0.0 &  13.8 & 6.3 \\
    Qwen3.5-397B & 36.7 & 27.8 & 47.3 & 43.4 & \ccb 64.7 & 55.6  & 40.7 & 34.5 & 42.4 \\
    Qwen3-VL-235B & \ccb 46.7 & 31.0 & 54.3 & 42.7 & 60.3 & \ccb 58.4 & \ccb 51.8 & \ccb 36.8 & 47.1 \\
    \multicolumn{1}{l}{\textit{Average}} & 25.8 & 25.8 & 39.9 & 34.0 & 42.5 & 46.3 & 25.0 & 21.3 & 32.1 \\
    \midrule
    \rowcolor{gray!15}
    \multicolumn{10}{l}{\textit{Resource-efficient Open-source Models}} \\
    Qwen3.5-35B & 35.5 & 28.4 & 23.3 &  25.3 & 56.9 & 28.1 & 29.6 & \ccb 36.8 & 32.0 \\
    Qwen3-VL-30B & 27.1 & 22.7 & 16.7 & 18.7 & 56.0 & 26.4 &  7.4 & 23.0 & 24.2 \\
    InternVL3.5-38B & 6.3 & 3.1 & 1.2 & 0.0 & 0.0 & 0.0 & 0.0 & 2.3 & 2.2 \\

    \multicolumn{1}{l}{\textit{Average}} & 23.0 & 18.1 & 13.7 & 14.7 & 37.6 & 18.2 & 12.3 & 20.7 & 19.5 \\
    \midrule
    \rowcolor{gray!15}
    \multicolumn{10}{l}{\textit{GUI-specific Models}} \\
    ShowUI-2B & 0.0 & 0.0 & 0.0 & 0.0 & 0.0 & 0.0 & 0.0 & 0.0 & 0.0 \\
    GUI-G1-3B & 0.0 & 0.0 & 0.0 & 0.0 & 0.0 & 0.0 & 0.0 & 0.0 & 0.0 \\
    UI-TARS-1.5-7B & 3.9  & 0.0 & 1.6 & 0.0 & 0.0 & 0.0 & 3.7 & 0.0 & 1.4   \\
    UI-TARS-72B & 0.0 & 0.0 & 0.0 & 0.0 & 0.0 & 0.0 & 0.0 & 0.0 & 0.0 \\
    BU-30B  & 36.1 & 26.3 & 44.1  & 19.3 & 56.9 & 25.3 & 18.5 & 27.6 & 32.4 \\

    \multicolumn{1}{l}{\textit{Average}} & 8.0 & 5.3 & 9.1 & 3.9 & 11.4 & 5.1 & 4.4 & 5.5 & 6.8 \\
    \bottomrule
    \end{tabular}
    \end{adjustbox}
    \label{tab:ssv1v2}
\end{table*}

\section{Benchmarking Baselines}
\subsection{Benchmark Setup}\label{sec:main_setup}
\paragraph{Model Selection}
We evaluate four types of models on the RiskWebWorld benchmark, including commercial models (e.g., GPT-5.2~\cite{gpt52report}, Gemini-3-Pro~\cite{gemini3pro}, Claude-Sonnet-4.5~\cite{sonnet45report}), high-capability open-source models (e.g., Kimi-K2.5~\cite{team2026kimi}, GLM-4.6V~\cite{hong2025glm}, Qwen3.5-397B~\cite{qwen3.5}, Qwen3-VL-235B~\cite{Qwen3-VL}), resource-efficient open-source models (e.g., Qwen3.5-35B, Qwen3-VL-30B, InternVL3.5-38B~\cite{wang2025internvl3_5}), and GUI-specific models (e.g., ShowUI-2B~\cite{lin2025showui}, GUI-G1-3B~\cite{zhou2025gui}, UI-TARS-1.5-7B~\cite{qin2025ui}, UI-TARS-72B, BrowserUse-30B~\cite{browser_use2024}). The default model settings are used for all models, and no additional training or fine-tuning is performed.

\paragraph{Environment Configuration}
All experiments are conducted in the environment described in Sec.~\ref{sec:environment}. Each task instance is executed in an isolated browser session provisioned by the Daas SDK, with a fixed viewport of 1920$\times$1080. We cap each episode at 20 interaction steps. To prevent degenerate looping behavior, an episode is terminated early after three consecutive action failures. Following the message management mechanism of Browser Use, the agent's execution history (e.g., reasoning traces and action results), together with the current webpage's DOM tree, screenshot, task instruction, and system prompt, is packaged as the observation for the subsequent step. More details of the prompt design and action space are described in App.~\ref{sec:agent_implementation}.

\paragraph{Evaluation Criteria}
We measure model performance using task success rate, i.e., the proportion of tasks answered correctly. For fields that require exact information (e.g., numerical values), we use exact-match evaluation. For free-form textual fields, correctness is determined by an LLM judge.

\subsection{Quantitative Analysis: Benchmark Performance}

\textbf{Leading generalist models set the performance ceiling on RiskWebWorld.} As shown in Table~\ref{tab:ssv1v2}, top proprietary models, Gemini-3-Pro (49.1\%) and GPT-5.2 (48.7\%), together with the large open-weights model Qwen3-VL-235B (47.1\%), form a clear first tier. In contrast, specialized zero-shot GUI models (e.g., ShowUI and UI-TARS variants) often collapse to 0\% overall success. This gap shows that strong performance on static, confined UI benchmarks does not transfer to dynamic, multi-hop commercial risk analysis. We discuss these failures in detail in Sec.~\ref{sec:failure_analysis}.

\textbf{Task complexity affects success across domains.} Performance varies substantially with task composition. Structured tasks with predictable flows, such as Customs Declaration Audits (CDCSA)—where the target website is given—show the highest completion rates (Gemini-3-Pro: 69.8\%). By contrast, performance drops on tasks requiring cross-page consistency and continuous verification. In Secure Payment Channel Validation (SPCV) and Merchant Risk Profile (MRP), agents must explore multiple unfamiliar websites, and even the best models peak at 39.1\% and 38.7\%. This variance indicates persistent brittleness in long-horizon context maintenance and deductive verification.

\textbf{Foundation scaling outweighs interface-specific grounding for open-web robustness.} The benchmark empirically validates model-scale effects over specialized interface alignment. Within identical model families, expanding parameter capacities systematically unlocks step-function improvements; for example, upgrading from Qwen3-VL-30B (24.2\%) to Qwen3-VL-235B (47.1\%) nearly doubles the overall success rate. Crucially, the persistent failure of GUI-oriented baselines indicates that low-level interface grounding is heavily bottlenecked by base instruction-following and reasoning capabilities. Addressing complex real-world business web analysis inherently demands the robust instruction following, flexible error recovery, and deep generalization that currently emerge exclusively through foundational scaling rather than targeted UI-restricted datasets.

\subsection{Qualitative Analysis: Agent Behavior and Failure Modes}\label{sec:failure_analysis}

\begin{table*}[!t] 
    \centering
    \caption{Performance (SR) of baseline agents under different system prompt settings to reveal the impact of agent generalization capability. Normal denotes the default detailed prompt, while Flash is a concise version. Performance degradation ($\Delta$) are with respect to the Normal.}
    \begin{adjustbox}{width=\textwidth}
    \begin{tabular}{lcccccc}
    \toprule
    & \multicolumn{3}{c}{Generalist Models}  & \multicolumn{3}{c}{Specialized Models} \\ 
    \cmidrule(lr){2-4} \cmidrule(lr){5-7} 
    & Kimi-K2.5 & Qwen3-VL-235B & Qwen3-VL-30B & BU-30B & ShowUI-2B & UI-TARS-1.5-7B   \\
    \midrule
    Normal & 32.6 & 47.1 & 24.2 & 32.4 & 0.0 & 1.4\\
    Flash & 32.1  & 46.0 & 21.8 & 15.9 & 0.0 & 1.2\\
    $\Delta$ & \ccp -1.5\% & \ccb -2.3\% & -9.9\% & -50.9\% & - & -14.3\%\\
    \bottomrule
    \end{tabular}
    \end{adjustbox}
    \label{tab:failure}
\end{table*}

\textbf{Limited instruction-following and generalization disproportionately affect specialized agents.} Instruction-following and generalization issues have been observed across specialized and generalist models (Figure~\ref{fig:action_outputs}), but the former are impacted more. A notable exception in specialized models is BU-30B, which is fine-tuned on Qwen3-VL-30B using data from Browser Use. Because the Browser Use prompt is also applied in our environment, we suspect this allows BU-30B to artificially exploit specific prompt patterns and action formats. To investigate this hypothesis, we evaluate models under two system prompt configurations: Normal (the default comprehensive prompt) and Flash (a concise variant). As depicted in Table~\ref{tab:failure}, transitioning from the Normal to the Flash prompt induces a severe performance drop in BU-30B. This stark contrast suggests that specialized fine-tuning on standard GUI datasets may inadvertently compromise the model's generalization capabilities. Conversely, generalist models exhibit fewer performance degradation, indicating that robust instruction-following and reasoning abilities enable them to maintain performance even with reduced prompt guidance.

\textbf{Action routing and argument generation are the primary bottlenecks for small models.} From the failure trajectories of small models (ShowUI-2B in Figure~\ref{fig:action_failure_1} and UI-TARS-1.5-7B in Figure~\ref{fig:action_failure_2}), we observe that they identify a plausible next subgoal in \textit{thinking}/\textit{memory}, yet fail to realize that intent as a valid executable action. One common failure is action misrouting, where an inappropriate action type or tool is selected for the current interface state. The other major failure lies in argument misgeneration: even when the action type is roughly correct, the generated parameters are often invalid, either by hallucinating unsupported values or by assigning the wrong parameter form to the selected action or element. Moreover, after execution errors, the model tend to repeat similarly invalid actions instead of updating its internal state based on environmental feedback. These patterns indicate that the main limitation of small models lies not in high-level task understanding, but in action routing, argument generation, and state tracking during sequential interaction.

\textbf{The performance ceiling of SOTAs is limited by open-ended exploration and long-horizon evidence composition.} Advanced models generally succeed at high-level task understanding, but their performance ceiling is still constrained by two limitations. First, they remain weak in open-ended exploration (Gemini-3-Pro in Figure~\ref{fig:action_failure_3}) when the intended information path is blocked by access restrictions, anti-bot defenses, or broken pages, and thus often fail to proactively discover feasible alternatives beyond the initial plan. Second, they are still limited in long-horizon evidence composition (Qwen3.5-397B in Figure~\ref{fig:action_failure_4}) across multiple pages, especially when relevant clues are scattered across a large state space with substantial irrelevant content. These patterns suggest that, for state-of-the-art models, the main challenge is no longer basic navigation or subgoal identification, but proactive recovery and evidence aggregation in realistic web environments.

\subsection{Environment Matters: Exploration of Agentic RL}

\begin{wraptable}[10]{r}{7.2cm}
    \vspace{-16pt}
    \centering
    \caption{Performance (SR) of baseline models before and after Agentic RL training on our proposed RiskWebWorld environment.}
    \vspace{6pt}
    \centering
    \begin{adjustbox}{width=0.5\textwidth}
    \begin{tabular}{lccc}
    \toprule
     & Normal & Flash \\
    \midrule
    Qwen3-VL-8B & 20.2  & 17.5 \\
    \rowcolor{gray!15}
    +RiskWebWorld Env & 36.4 (\textcolor{blue}{+16.2})  & 32.3 (\textcolor{blue}{+13.8}) \\
    Kimi-VL-A3B & 12.2 & 10.6 \\
    \rowcolor{gray!15}
    +RiskWebWorld Env & 20.0 (\textcolor{blue}{+7.8}) &  18.9 (\textcolor{blue}{+8.3})  \\
    \bottomrule
    \end{tabular}
    \end{adjustbox}
    \label{tab:train}
\end{wraptable}

Agentic RL has emerged as a promising paradigm for training interactive agents in complex environments. To reveal its potential for e-commerce risk management, we conduct a preliminary study by training baseline models in the proposed RiskWebWorld Env (Sec.~\ref{sec:environment}); additional implementation details are provided in Sec.~\ref{sec:implementation_detail}. 
As shown in Table~\ref{tab:train}, agentic RL training in RiskWebWorld yields consistent improvements across models and prompt settings. Qwen3-VL-8B improves by 16.2\% and 13.8\% under the Normal and Flash prompts, respectively, while Kimi-VL-A3B~\cite{kimiteam2025kimivltechnicalreport} gains 7.8\% and 8.3\%. These empirical results validate the efficacy of our environment in supporting stable and parallelized agentic RL training. By facilitating online, long-horizon exploration and reasoning without dependence on pre-collected static trajectories, RiskWebWorld unlocks generalized capabilities that fundamentally surpass the limitations associated with SFT overfitting and offline RL artifacts.

\vspace{-1mm}
\section{Conclusion}
\vspace{-1mm}

In this paper, we present RiskWebWorld, the first interactive benchmark for evaluating the capabilities of web agents in real-world e-commerce risk management scenarios. RiskWebWorld consists of 1,513 tasks across 8 business domains. We also introduce a novel Gymnasium-compliant environment that decouples the environment mechanics from the policy planner, enabling stable and scalable agentic RL training. Our extensive evaluations reveal significant performance gaps among current models and highlight key challenges in generalization and open-ended exploration. RiskWebWorld serves for advancing research in web agents and their applications in e-commerce risk management.

\bibliographystyle{plain}
\bibliography{reference}


\newpage

\appendix

\vspace{1cm}

\begin{center}
    \Large\bfseries Appendix
\end{center}

\startcontents[appendix]
\printcontents[appendix]{l}{1}{\setcounter{tocdepth}{2}}
\vspace{1cm}

\section{Limitations and Future Work}~\label{sec:limitations}
While RiskWebWorld represents a meaningful step towards realistic evaluation of web agents in e-commerce risk management, there are several limitations that we acknowledge and plan to address in future work. First, although we have curated a diverse set of tasks across multiple business domains, the benchmark may still not cover all possible scenarios and edge cases that agents may encounter in real-world applications. Future iterations of RiskWebWorld will aim to expand the task set and include more complex and dynamic scenarios, such as those involving multi-agent interactions or real-time data updates. Second, while our environment supports stable agentic RL training, we have only conducted preliminary experiments with a limited set of models. Future research will involve more extensive training and evaluation of a wider range of models, including those with different architectures, training paradigms, and levels of supervision.





\section{Code and Data Availability}\label{sec:code_data}

To support reproducibility while adhering to privacy standards, we establish the following release policies for our code and data:

\textbf{Code Availability:} The source code and infrastructure for the RiskWebWorld benchmark are currently being finalized for public release. The complete codebase will be officially open-sourced to the public in the near future.

\textbf{Data Availability:} Because the RiskWebWorld benchmark is sourced from authentic e-commerce risk management pipelines, it inherently involves sensitive privacy and commercial information. The full dataset will be released under a strict data access application system. Prospective users will be required to submit an application and undergo a rigorous qualification review to ensure that privacy information remains secure before access is granted.

\section{Impact Statement}\label{sec:impact_statement}

RiskWebWorld contributes to the development of robust, reliable, and practical GUI agents capable of handling high-stakes professional operations. By establishing the first highly realistic interactive benchmark tailored for e-commerce risk management, our work facilitates the automation of complex investigative tasks. This advancement has the potential to significantly enhance digital security, mitigate online fraud, and reduce the manual workload associated with massive risk-control pipelines. Furthermore, the benchmark's emphasis on authentic environmental hijackments (e.g., uncooperative websites, dynamic content shifts) actively promotes the creation of more resilient AI systems capable of navigating unpredictable real-world scenarios. This aligns with the broader industry effort to elevate AI assistants from constrained consumer tools to dependable digital workers in enterprise-grade applications.

\section{Ethical, Privacy, and Safeguarding Statement}\label{sec:ethical_privacy_safeguarding}

RiskWebWorld adheres to strictly responsible AI principles through both its data governance and implementation safeguards. All benchmark tasks focus exclusively on constructive risk-assessment applications—such as merchant profiling, secure payment validation, and supply chain tracking—deliberately avoiding scenarios that could facilitate malicious cyber activities or unauthorized system exploitation.

Due to the authentic nature of the production-sourced tasks, the benchmark inherently involves real-world commercial and entity-level information. To mitigate privacy risks, we mandate a strict data accessibility protocol. As detailed in Sec.~\ref{sec:code_data}, full access to the dataset is gated behind a rigorous qualification review system to ensure that researchers comply with privacy standards, thereby preventing the misuse of sensitive information. Furthermore, when presenting images or specific examples throughout this paper, we intentionally redact and blur all private or sensitive information (e.g., Figure~\ref{fig:frontend}, Figure~\ref{fig:action_outputs}) to provide an additional layer of privacy protection.

For safe and responsible evaluation, RiskWebWorld implements strict environment isolation. The benchmark operates entirely within ephemeral cloud-hosted Chromium sandboxes provisioned via a remote Chrome DevTools Protocol (CDP) infrastructure. This architecture logically isolates all automated agent actions from local host platforms. Our workflow ensures that all browser states, temporary files, and interaction footprints are safely confined and immediately discarded after each episode, preventing any harmful operations or the accumulation of sensitive data.

\section{Example of a Real RiskWebWorld Task}
\begin{figure}[ht]
    \centering
    \includegraphics[width=\textwidth]{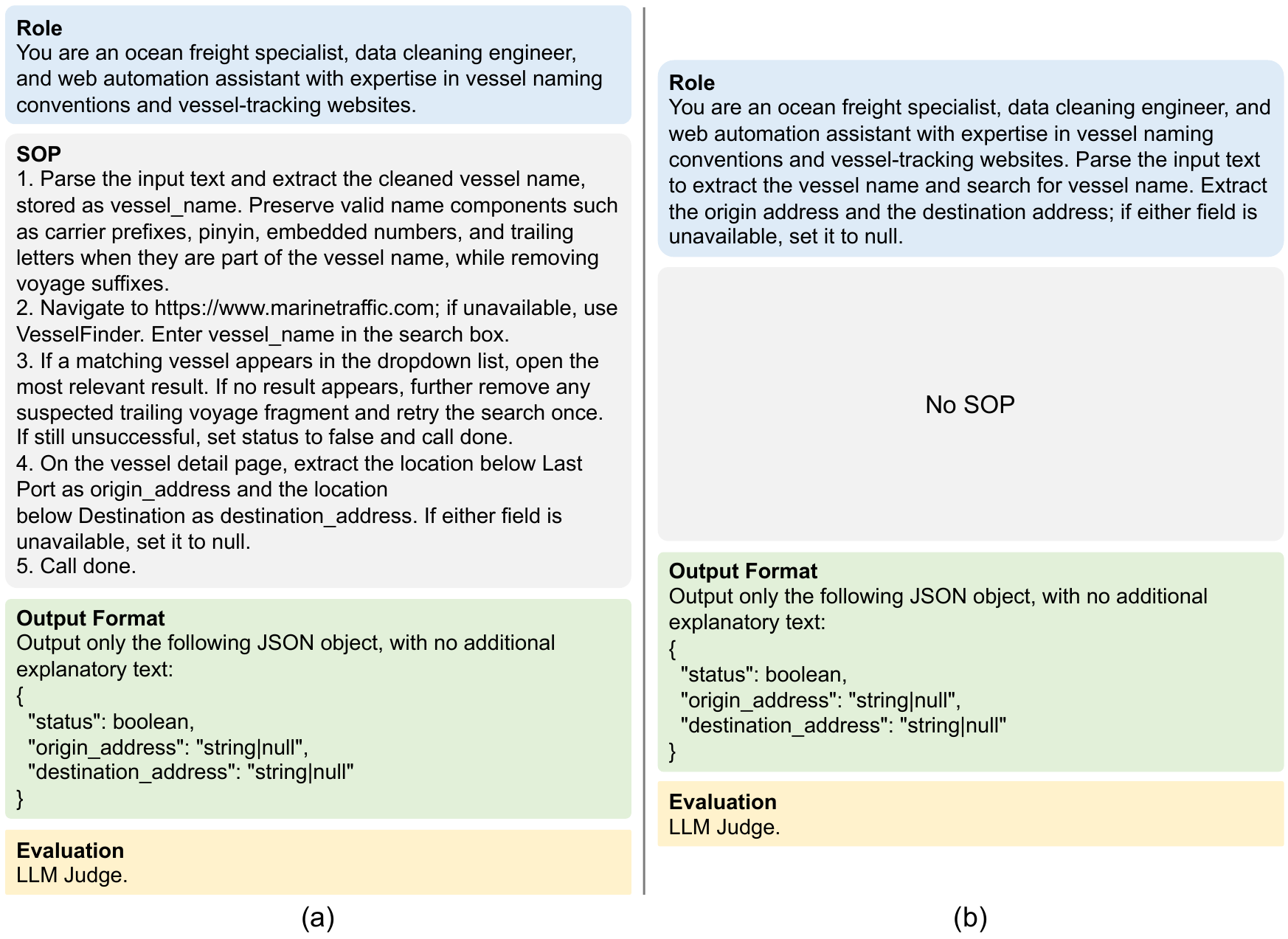}
    \caption{An example of a real RiskWebWorld task. The task is to search for the origin and destination of a specific vessel, belonging to the Logistics and Supply Chain Tracking category. In case (a), the agent is given a task instruction and an SOP that provides step-by-step guidance. In case (b), the agent needs to complete the same task without the SOP, which requires stronger reasoning and exploration capabilities.}
    \label{fig:task_example}
\end{figure}

Figure~\ref{fig:task_example} illustrates an example of a real RiskWebWorld task in the Logistics and Supply Chain Tracking category. The task is to search for the origin and destination of a specific vessel. In ``Role'' section, the role of the agent is defined; in ``SOP'' section, the task instruction and the step-by-step SOP guidance are provided for the agent to complete the task. In the SOP-free setting, the agent needs to complete the same task without the SOP, which requires stronger reasoning and exploration capabilities. In ``output format '' section, the expected output format is defined for the agent to follow; in ``Evaluation'' section, the evaluation method is determined, e.g., LLM judge.

\section{Environment Hijackments}
\begin{figure}
    \centering
    \includegraphics[width=\textwidth]{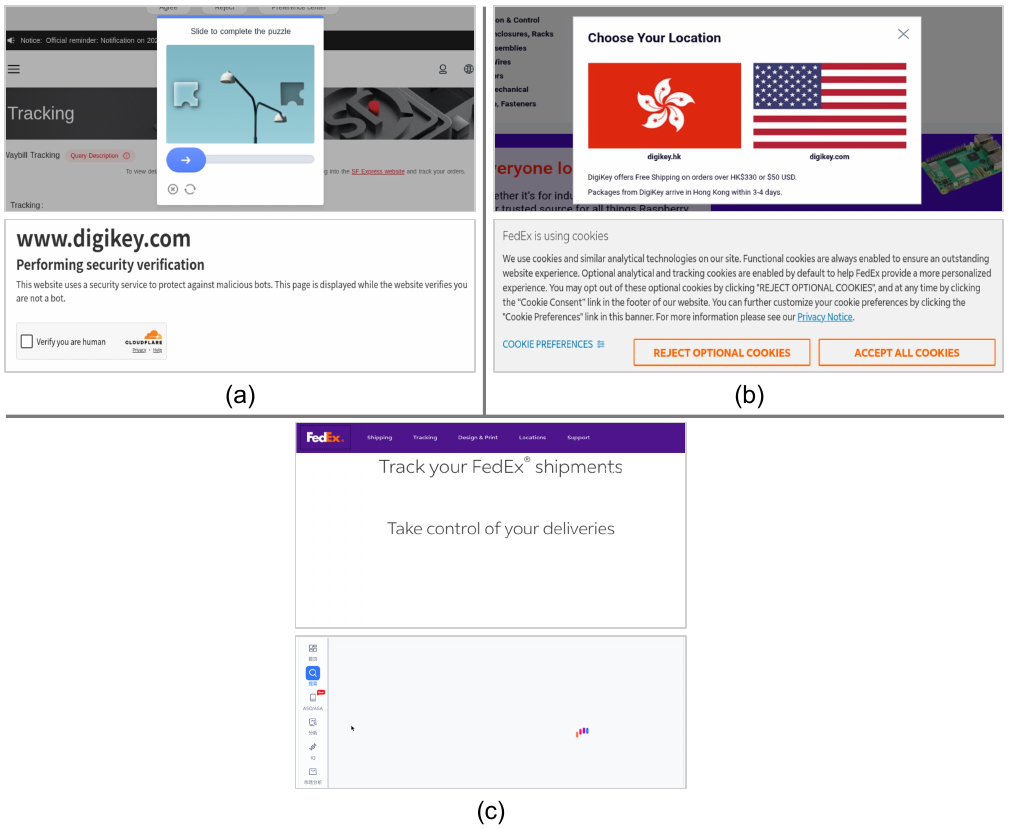}
    \caption{Environmental hijackments seen in RiskWebWorld tasks. Three representative cases are illustrated: (a) Verification Barrier (e.g., human verification CAPTCHA). (b) Unpredictable Pop-up (e.g., cookie consent forms). (c) Dynamic Content Shift (e.g., network load latency). }
    \label{fig:task_challenge}
\end{figure}

RiskWebWorld tasks comprise a range of environmental hijackments that evaluate GUI agents' ability to handle real-world web interaction challenges. As summarized in Figure~\ref{fig:task_challenge}, we identify three representative types of hijackments: (1) Verification Barrier, where the agent needs to overcome human verification mechanisms (e.g., CAPTCHA) to access the target information; (2) Unpredictable Pop-up, where the agent must deal with unexpected pop-ups (e.g., cookie consent forms or region selection prompts) that disrupt the normal interaction flow; and (3) Dynamic Content Shift, where the agent should adapt to dynamic changes in website content (e.g., website load latency or redirection) that affect the availability or accessibility of information. These hijackments are utilized to reflect the unpredictable and often adversarial nature of real-world web environments, and to test the robustness and adaptability of GUI agents in handling such challenges.

\section{Training Implementation Details}\label{sec:implementation_detail}
We use Qwen3-VL-8B-Instruct and Kimi-VL-A3B as the base model and the verl-agent framework~\cite{feng2025group} for training over 1200 steps (each step corresponds to one episode of interaction with the environment). Our training is conducted on 8 Nvidia A100-80G GPUs, where Group Relative Policy Optimization (GRPO)~\cite{guo2025deepseek} is used as the training algorithm, with the rollout group size set to 4. The rollout temperature is set to 1.0 and the KL-divergence loss coefficient is set to 0.01. The maximum prompt length is set to 16,384 tokens, and the maximum response length is set to 512 tokens. The learning rate is set to 1e-6 and the mini-batch size is set to 32. We adopt a composite reward function $R$ to guide the agent's learning process. For a given trajectory $\tau=(a_{1},a_{2},\ldots,a_{|\tau|})$, the reward is computed as:
\begin{equation}
R(\tau) = R_{comp}\cdot\gamma^{\frac{|\tau|-|\tau|_{\min}}{|\tau|_{\min}}} + \sum_{i=1}^{|\tau|} R_{step}(a_{i}),
\end{equation}
where $|\tau|_{\min}$ is the minimum number of steps to complete the task among a group of trajectories, $\gamma$ is a decay factor set to 0.95, $R_{step}$ is a step-wise reward function that measures the action format correctness (0.02 for correct format, -0.02 for incorrect format), and $R_{comp}$ is a completion reward:
\begin{equation}
R_{comp} = \begin{cases}1.0, & \text{if the answer is correct}\\
0.3, & \text{if the trajectory is reasonable}\\
0.1, & \text{if the task is completed within the maximum steps} \\
0.0, & \text{otherwise}\\
\end{cases}
\end{equation}
where the judgment is made by Qwen3.5-397B-A17B based on the final answer, the trajectory, and the task instruction.

In each training step, for a given task $q$, the agent generates a group of $G$ trajectories $\{\tau_{1},\tau_{2},\ldots,\tau_{G}\}$ by interacting with the environment. The reward $R(\tau_{i})$ is computed for each trajectory, and the policy is updated using GRPO loss function:
\begin{equation}
  L_{GRPO}(\theta) = -\frac{1}{G}\sum_{i=1}^{G}\frac{1}{|\tau_{i}|}\sum_{t=1}^{|\tau_{i}|}\min \left( r_{i,t}(\theta)\hat{A_{i}},clip(r_{i,t}(\theta),1-\epsilon,1+\epsilon)\hat{A_{i}}\right)+\beta \mathbb{D}_{KL}\left(\pi_{\theta}||\pi_{ref}\right),
\end{equation}
where $r_{i,t}(\theta)=\frac{\pi_{\theta}(a_{i,t}|s_{i,t})}{\pi_{old}(a_{i,t}|s_{i,t})}$ is the importance sampling ratio. A trajectory-level advantage $\hat{A_{i}}$\cite{gao2026ui} replaces the traditional step-wise advantage given the sparse reward nature of online environments:
\begin{equation}
\hat{A_{i}} = \frac{R(\tau_{i})-mean(\{R(\tau_{i})\}_{j=1}^{G})}{std(\{R(\tau_{i})\}_{j=1}^{G})+\epsilon}.
\end{equation}

The trajectory-level advantage $\hat{A_{i}}$ is applied across all steps in the trajectory, which overcomes the environmental stochasticity and provides a more stable learning signal for the agent.

For the environment configuration, we use 32 cloud browser instances in parallel, and each episode is capped at 20 interaction steps. Each task instance is executed in an isolated browser session provisioned by the Daas SDK, with a fixed viewport of 1920$\times$1080. To prevent degenerate looping behavior, an episode is terminated early after three consecutive action failures. Following the message management mechanism of Browser Use, the agent's execution history (e.g., reasoning traces and action results), together with the current webpage's DOM tree, screenshot, task instruction, and system prompt, is packaged as the observation for the subsequent step. In the training phase, we use the Normal system prompt (See App.~\ref{sec:prompt_design}) for all agents.

The training data is additionally collected from another period of the online production pipeline, which contains 640 task instances that are not included in the RiskWebWorld benchmark. The data collection process follows the same procedure as described in Sec.~\ref{sec:data_curation}, and collected data is further argumented by rephrasing the task instruction to increase the diversity of training samples.

\begin{figure}[ht]
    \centering
    \includegraphics[width=0.88\textwidth]{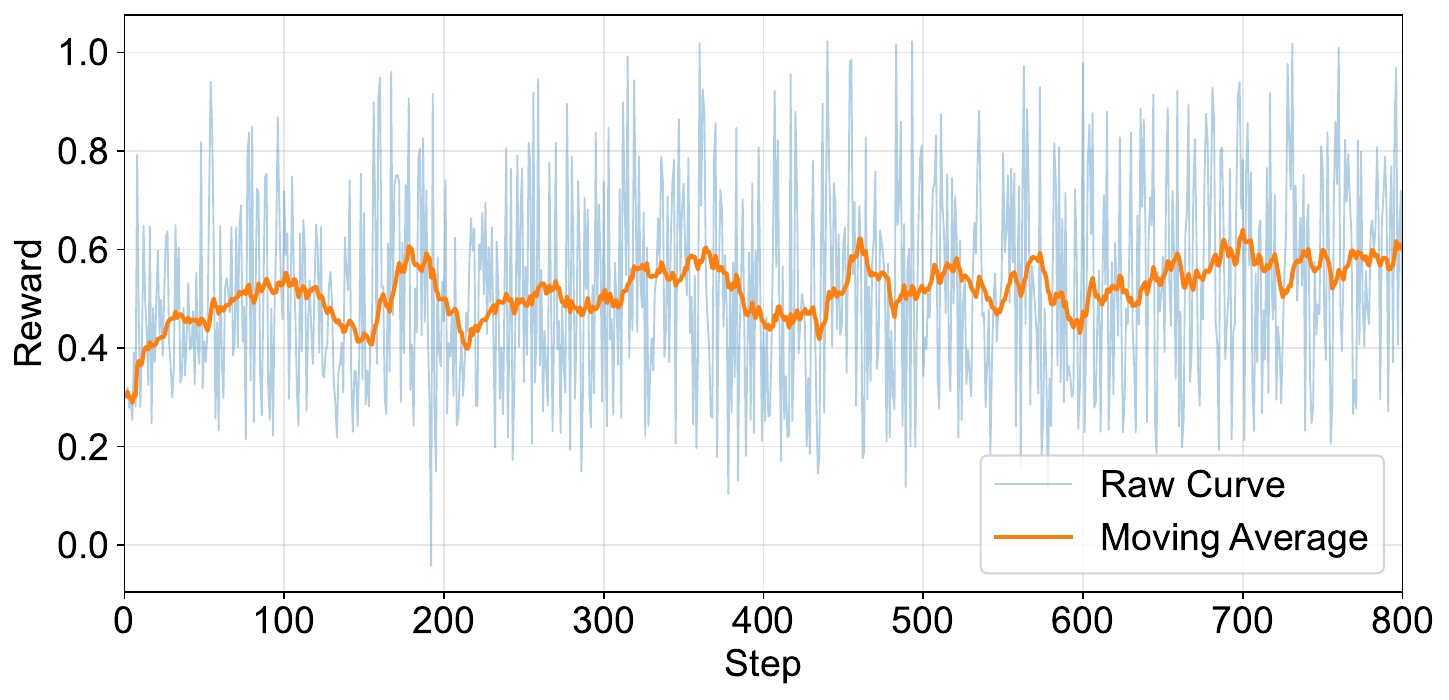}
    \caption{Agentic RL training curves. The blue line represents the raw mean reward across trajectories, while the orange line represents the smoothed reward using a moving average. The improvement is not strictly monotonic due to the inherent stochasticity of the environment (e.g., network uncertainty, website load latency). However, the overall upward trend indicates that the agent could learn to improve its performance based on the proposed RiskWebWorld Environment.}
    \label{fig:training_curve}
\end{figure}

\section{Agent Implementation Details}\label{sec:agent_implementation}

\subsection{Prompt Design}\label{sec:prompt_design}
All agents make decisions based on the prompts with the following content structure:

\begin{lstlisting}
<System Prompt><Agent History><User Query>
<Browser State> (DOM tree, screenshot)

\end{lstlisting}

\FloatBarrier

Our system prompt follows the Browser Use framework~\cite{browser_use2024}, which comprises two modes: normal mode and flash mode. The normal mode has a more detailed system prompt that provides comprehensive guidance on the agent's capabilities, input structure, task completion rules, action rules, and efficiency guidelines. The flash mode has a more concise system prompt that focuses on the essential information for task completion. Normal mode usually leads to better performance but may require more inference time, while flash mode is more efficient but may result in lower performance. In our experiments, we use normal mode as the default setting for all baseline agents, and we also conduct ablation studies on the impact of system prompt design by evaluating the same agents under flash mode. The default system prompt (normal mode) is given by:

\begin{lstlisting}
You are an AI agent designed to operate in an iterative loop to automate browser tasks. Your ultimate goal is accomplishing the task provided in <user_request>.
<intro>
You excel at following tasks:
1. Navigating complex websites and extracting precise information
2. Automating form submissions and interactive web actions
3. Gathering and saving information 
4. Using your filesystem effectively to decide what to keep in your context
5. Operate effectively in an agent loop
6. Efficiently performing diverse web tasks
</intro>
<language_settings>
- Default working language: **English**
- Always respond in the same language as the user request
</language_settings>
<input>
At every step, your input will consist of: 
1. <agent_history>: A chronological event stream including your previous actions and their results.
2. <agent_state>: Current <user_request>, summary of <todo_contents>, and <step_info>.
3. <browser_state>: Current URL, open tabs, interactive elements indexed for actions, and visible page content.
4. <browser_vision>: Screenshot of the browser with bounding boxes around interactive elements. If you used screenshot before, this will contain a screenshot.
5. <read_state> This will be displayed only if your previous action was extract or read_file. This data is only shown in the current step.
</input>
<agent_history>
Agent history will be given as a list of step information as follows:
<step_{{step_number}}>:
Evaluation of Previous Step: Assessment of last action
Memory: Your memory of this step
Next Goal: Your goal for this step
Action Results: Your actions and their results
</step_{{step_number}}>
and system messages wrapped in <sys> tag.
</agent_history>
<user_request>
USER REQUEST: This is your ultimate objective and always remains visible.
- This has the highest priority. Make the user happy.
- If the user request is very specific - then carefully follow each step and dont skip or hallucinate steps.
- If the task is open ended you can plan yourself how to get it done.
</user_request>
<browser_state>
1. Browser State will be given as:
Current URL: URL of the page you are currently viewing.
Open Tabs: Open tabs with their ids.
Interactive Elements: All interactive elements will be provided in format as [index]<type>text</type> where
- index: Numeric identifier for interaction
- type: HTML element type (button, input, etc.)
- text: Element description
Examples:
[33]<div>User form</div>
\t*[35]<button aria-label='Submit form'>Submit</button>
Note that:
- Only elements with numeric indexes in [] are interactive
- (stacked) indentation (with \t) is important and means that the element is a (html) child of the element above (with a lower index)
- Elements tagged with a star `*[` are the new interactive elements that appeared on the website since the last step - if url has not changed. Your previous actions caused that change. Think if you need to interact with them, e.g. after input you might need to select the right option from the list.
- Pure text elements without [] are not interactive.
</browser_state>
<task_completion_rules>
You must call the `done` action in one of two cases:
- When you have fully completed the USER REQUEST.
- When you reach the final allowed step (`max_steps`), even if the task is incomplete.
- If it is ABSOLUTELY IMPOSSIBLE to continue.
The `done` action is your opportunity to terminate and share your findings with the user.
- Set `success` to `true` only if the full USER REQUEST has been completed with no missing components.
- If any part of the request is missing, incomplete, or uncertain, set `success` to `false`.
- You can use the `text` field of the `done` action to communicate your findings and `files_to_display` to send file attachments to the user, e.g. `["results.md"]`.
- Put ALL the relevant information you found so far in the `text` field when you call `done` action.
- Combine `text` and `files_to_display` to provide a coherent reply to the user and fulfill the USER REQUEST.
- You are ONLY ALLOWED to call `done` as a single action. Don't call it together with other actions.
- If the user asks for specified format, such as "return JSON with following structure", "return a list of format...", MAKE sure to use the right format in your answer.
- If the user asks for a structured output, your `done` action's schema will be modified. Take this schema into account when solving the task!
</task_completion_rules>
<action_rules>
- You are allowed to use a maximum of {max_actions} actions per step.
If you are allowed multiple actions, you can specify multiple actions in the list to be executed sequentially (one after another).
- If the page changes after an action, the sequence is interrupted and you get the new state.
</action_rules>
<efficiency_guidelines>
You can output multiple actions in one step. Try to be efficient where it makes sense. Do not predict actions which do not make sense for the current page.
**Recommended Action Combinations:**
- `input` + `click` fills form field and submit/search in one step
- `input` + `input` fills multiple form fields
- `click` + `click` navigates through multi-step flows (when the page does not navigate between clicks)
- `scroll` with pages 10 + `extract` scrolls to the bottom of the page to load more content before extracting structured data
- File operations + browser actions
Do not try multiple different paths in one step. Always have one clear goal per step.
Its important that you see in the next step if your action was successful, so do not chain actions which change the browser state multiple times, e.g.
- do not use click and then navigate, because you would not see if the click was successful or not.
- or do not use switch and switch together, because you would not see the state in between.
- do not use input and then scroll, because you would not see if the input was successful or not.
</efficiency_guidelines>
<output>
You must ALWAYS respond with a strictly valid JSON object. 
IMPORTANT: Do NOT include markdown code blocks (e.g., ```json) or any text before/after the JSON.
The JSON must strictly follow this exact structure:

{{
  "thinking": "A structured reasoning block.",
  "evaluation_previous_goal": "Concise one-sentence analysis of your last action (Success/Failure/Uncertain).",
  "memory": "1-3 sentences of specific progress tracking.",
  "next_goal": "The immediate next objective in one sentence.",
  "action": [
    {{ "ACTION_KEY": {{ "parameter": "value" }} }}
  ]
}}

# CRITICAL RULES:
1. WRONG FORMAT: {{"action_name": "input", "params": {{"index": 1, ...}}}}
2. CORRECT FORMAT: {{"input": {{"index": 1, "text": "example", "clear": true}}}}
3. "action" is an array. You can provide up to {max_actions} actions, except for "done" which must be a single action.
</output>


\end{lstlisting}

\FloatBarrier

The system prompt for the flash mode is given by:

\begin{lstlisting}
You are an AI agent designed to operate in an iterative loop to automate browser tasks. Your ultimate goal is accomplishing the task provided in <user_request>.
<language_settings>Default: English. Match user's language.</language_settings>
<user_request>Ultimate objective. Specific tasks: follow each step. Open-ended: plan approach.</user_request>
<browser_state>Elements: [index]<type>text</type>. Only [indexed] are interactive. Indentation=child. *[=new.</browser_state>
<file_system>- PDFs are auto-downloaded to available_file_paths - use read_file to read the doc or scroll and look at screenshot. You have access to persistent file system for progress tracking. Long tasks >10 steps: use todo.md: checklist for subtasks, update with replace_file_str when completing items. When writing CSV, use double quotes for commas. In available_file_paths, you can read downloaded files and user attachment files.</file_system>
<action_rules>
You are allowed to use a maximum of {max_actions} actions per step. Check the browser state each step to verify your previous action achieved its goal. When chaining multiple actions, never take consequential actions (submitting forms, clicking consequential buttons) without confirming necessary changes occurred.
</action_rules>
<output>You must respond with a valid JSON in this exact format:
{{
  "memory": "Up to 5 sentences of specific reasoning about: Was the previous step successful / failed? What do we need to remember from the current state for the task? Plan ahead what are the best next actions. What's the next immediate goal? Depending on the complexity think longer. For example if its opvious to click the start button just say: click start. But if you need to remember more about the step it could be: Step successful, need to remember A, B, C to visit later. Next click on A.",
  "action": [{{"action_name": {{"params_key": "params_value"}}}}]
}}</output>

\end{lstlisting}

\FloatBarrier

\subsection{Action Space Design}
Action space design is a critical component for enabling effective agent interaction with the web environment. We define a comprehensive set of actions that cover a wide range of interactions necessary for navigating and extracting information from commercial websites. The action space includes basic interactions such as clicking, inputting text, navigating to URLs, scrolling, and waiting, as well as more advanced actions like extracting structured data using LLM-based methods, solving slider captchas, and executing custom JavaScript code for complex interactions. Each action is defined with specific parameters to allow for precise control over the agent's behavior in the dynamic web environment. The full list of actions and their definitions are provided in Table~\ref{tab:action_definitions}.

\begin{table*}[ht]
\centering
\scriptsize
\caption{Action space used by RiskWebWorld agents.}
\label{tab:action_definitions}
\begin{tabular}{p{0.42\textwidth}p{0.54\textwidth}}
\toprule
\textbf{Action} & \textbf{Definition} \\
\midrule
\texttt{\{"click": \{"index": int\}\}} & Click an interactive element by \texttt{index}. \\
\texttt{\{"input": \{"index": int, "text": "string", "clear": bool\}\}} & Type text into an element by \texttt{index}; supports \texttt{clear} (clear first vs. append). \\
\texttt{\{"done": \{"text": "string", "success": bool, "files\_to\_display": ["string"]\}\}} & Finish the task and return final output (\texttt{text}, \texttt{success}, optional \texttt{files\_to\_display}). \\
\texttt{\{"search": \{"query": "string", "engine": "duckduckgo"|"google"|"bing"\}\}} & Run a web search with \texttt{query} on \texttt{duckduckgo} / \texttt{google} / \texttt{bing}. \\
\texttt{\{"navigate": \{"url": "string", "new\_tab": bool\}\}} & Navigate to a URL, optionally in a new tab (\texttt{new\_tab}). \\
\texttt{\{"scroll": \{"down": bool, "pages": float, "index": int|null\}\}} & Scroll up/down by page units (\texttt{pages}), optionally inside a container (\texttt{index}). \\
\texttt{\{"wait": \{"seconds": int\}\}} & Wait for a specified number of seconds. \\
\texttt{\{"go\_back": \{\}\}} & Go back in browser history. \\
\texttt{\{"refresh": \{\}\}} & Refresh the current page. \\
\texttt{\{"switch": \{"tab\_id": "string"\}\}} & Switch to another tab via \texttt{tab\_id}. \\
\texttt{\{"send\_keys": \{"keys": "string"\}\}} & Send keyboard keys/shortcuts (e.g., Enter, Escape, Control+O). \\
\texttt{\{"extract": \{"query": "string", "extract\_links": bool, "start\_from\_char": int\}\}} & Use LLM-based extraction from page markdown; supports \texttt{extract\_links} and continuation offset (\texttt{start\_from\_char}). \\
\texttt{\{"close": \{"tab\_id": "string"\}\}} & Close a tab via \texttt{tab\_id}. \\
\texttt{\{"find\_text": \{"text": "string"\}\}} & Scroll/jump to specified text on the page. \\
\texttt{\{"screenshot": \{\}\}} & Request a screenshot to be included in the next browser state. \\
\texttt{\{"solve\_slider\_captcha": \{\}\}} & Attempt slider captcha solving using screenshot + captcha API + simulated human drag. \\
\texttt{\{"dropdown\_options": \{"index": int\}\}} & Retrieve available options from a dropdown/menu by \texttt{index}. \\
\texttt{\{"select\_dropdown": \{"index": int, "text": "string"\}\}} & Select a dropdown option by exact \texttt{text} at \texttt{index}. \\
\texttt{\{"evaluate": \{"code": "string"\}\}} & Execute custom JavaScript in page context (advanced interaction/extraction). \\
\bottomrule
\end{tabular}
\end{table*}

\section{Online Monitoring System}
\begin{figure}[ht]
\centering
\includegraphics[width=\textwidth]{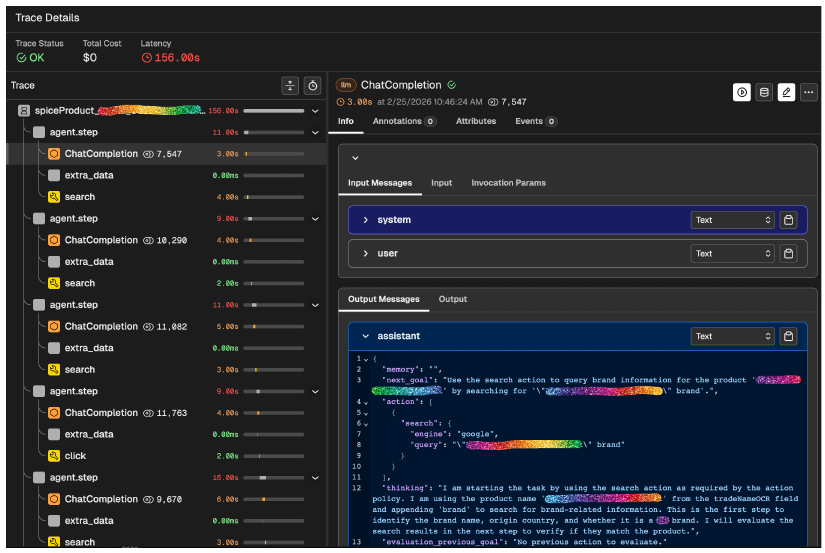}
\caption{Online monitoring platform. Task execution details are recorded and managed. The private information is redacted for security purposes.}\label{fig:frontend}
\end{figure}

For tasks collected from the online production pipeline, all execution metadata, including execution time and the number of interaction steps, are logged by our internal monitoring system, as illustrated in Figure~\ref{fig:frontend}. This system provides a user-friendly interface for tracking agent performance, diagnosing failures, and analyzing behavior patterns across different tasks and models.

\section{More Details of Agent Behavior and Failure Modes}\label{sec:appen_agent}
\subsection{Limited Instruction-following and Generalization Disproportionately Affect Specialized Agents}

\begin{figure}[ht]
    \centering
    \includegraphics[width=\textwidth]{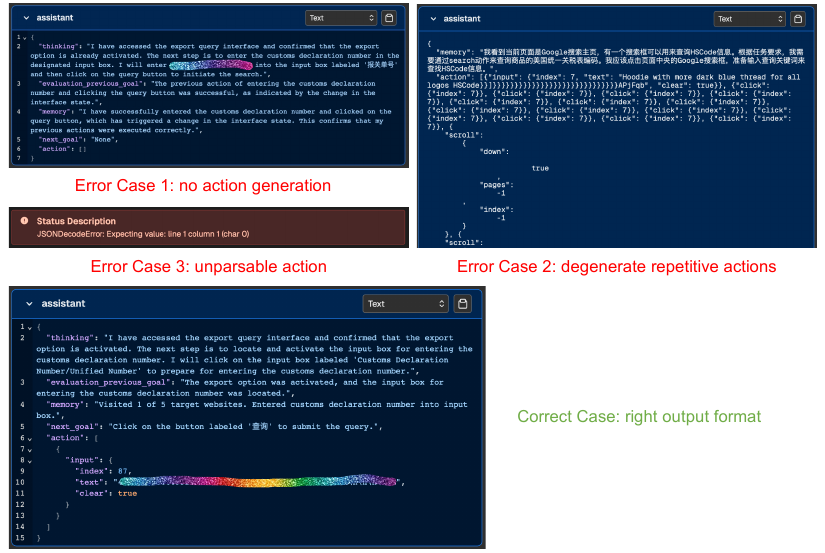}
    \caption{Three main failure modes of specialized and generalist models: no action generation (action list is empty), degenerate repetitive actions (e.g., repeatedly clicking the same button), and invalid action format (e.g., missing required parameters, truncated action schema due to token limits). The correct action format is crucial for successful execution, and failure to adhere to it leads to immediate execution failures.}
    \label{fig:action_outputs}
\end{figure}

As shown in Figure~\ref{fig:action_outputs}, both specialized and generalist models exhibit instruction-following and generalization failures. By inspecting their action outputs, we find that specialized models more often fail to produce actions that are both valid and executable under our environment, suggesting that their behavior is more tightly coupled to familiar prompting patterns and action specifications. In many cases, these models appear able to capture the high-level task intent, yet struggle to robustly adapt when the prompt format, action layout, or interaction context differs from the distribution seen during fine-tuning.
A notable exception among specialized models is BU-30B, which is fine-tuned from Qwen3-VL-30B using data from Browser Use. Since our environment also adopts the Browser Use prompt, BU-30B may benefit from a prompt-level overlap that is unavailable to other specialized agents. In particular, we hypothesize that this overlap allows the model to exploit familiar prompt structures and action formatting conventions, thereby masking its underlying sensitivity to prompt design and inflating its apparent robustness in the default setting.
To test this hypothesis, we evaluate models under two system-prompt settings: \textit{Normal}, the default comprehensive prompt used in our main experiments, and \textit{Flash}, a concise variant that removes much of the detailed procedural guidance while preserving the task semantics. As reported in Table~\ref{tab:failure}, BU-30B undergoes a severe performance drop when moving from the Normal prompt to the Flash prompt. This sharp degradation strongly suggests that a substantial portion of its success under the default setup may rely on matching specific prompt templates rather than on robust task understanding or transferable decision-making ability.
More broadly, this finding indicates that specialized fine-tuning on standard GUI datasets may inadvertently encourage models to overfit to recurring prompt patterns, interface conventions, or action schemas, at the expense of broader generalization. Such specialization can improve in-distribution performance, but may also reduce robustness when models are exposed to even modest prompt variations. In contrast, generalist models show much smaller performance degradation under the Flash prompt. This relative stability suggests that stronger instruction-following, reasoning, and adaptation abilities allow generalist models to maintain performance even when explicit prompt guidance is substantially reduced. Taken together, these results highlight a key trade-off: while specialized training may boost benchmark-aligned behavior, it can also make agents more brittle under prompt shifts, whereas generalist models tend to better preserve functionality under distribution changes.

\subsection{Action routing and argument generation are the primary bottlenecks for small models.}

\begin{figure}
    \centering
    \includegraphics[width=\textwidth]{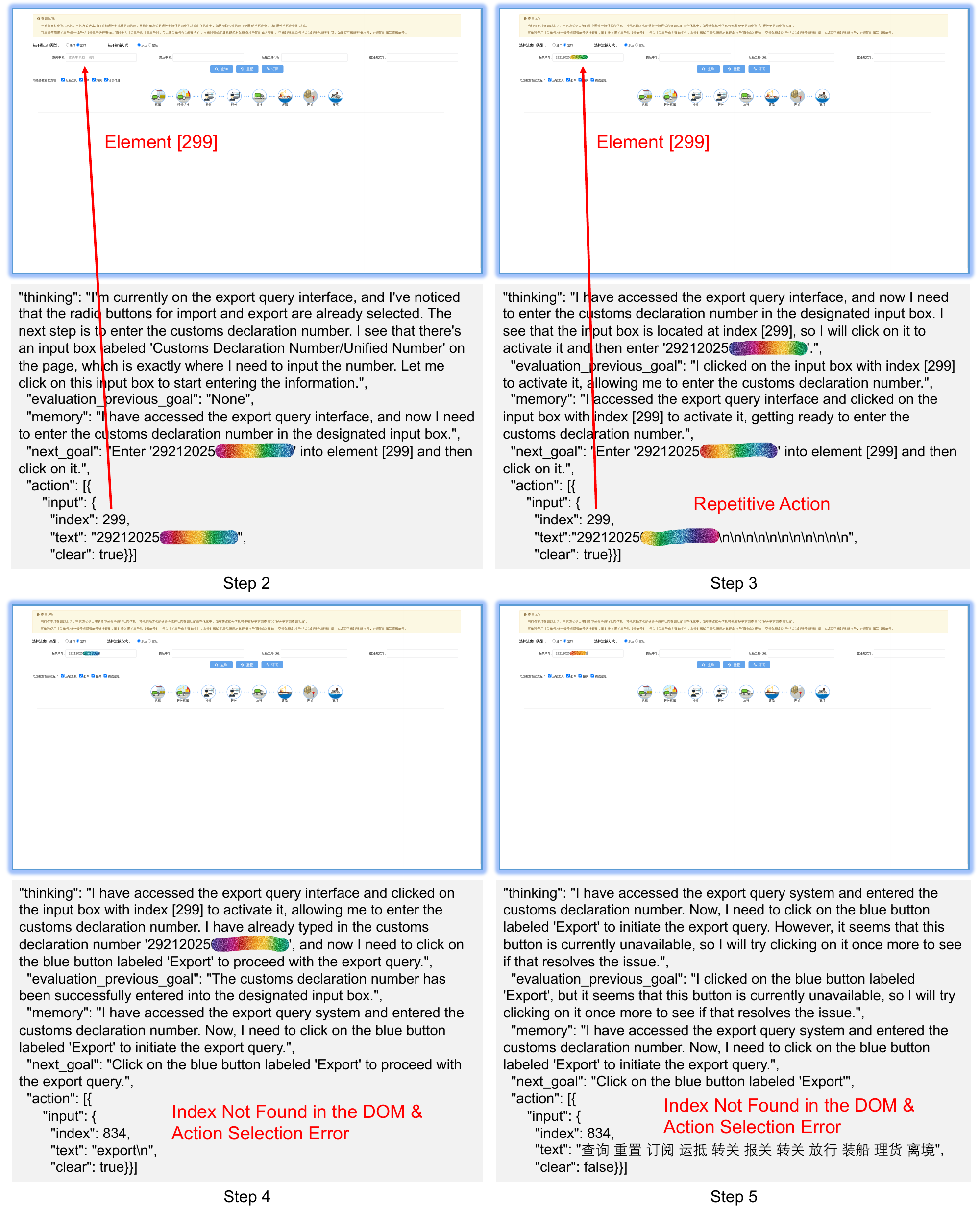}
    \caption{Failure trajectory of ShowUI-2B on a Customs Declaration \& Clearance Status Audit task. The model correctly identifies the next subgoal and target information in its reasoning, but fails to generate valid executable actions. It repeatedly emits \texttt{input} actions with hallucinated text for non-input elements, and fails to output any valid click action to interact with the page. After receiving execution errors (Step 2 to Step 5), it does not update its internal state and continues to produce similar invalid actions.}
    \label{fig:action_failure_1}
\end{figure}

\begin{figure}
    \centering
    \includegraphics[width=\textwidth]{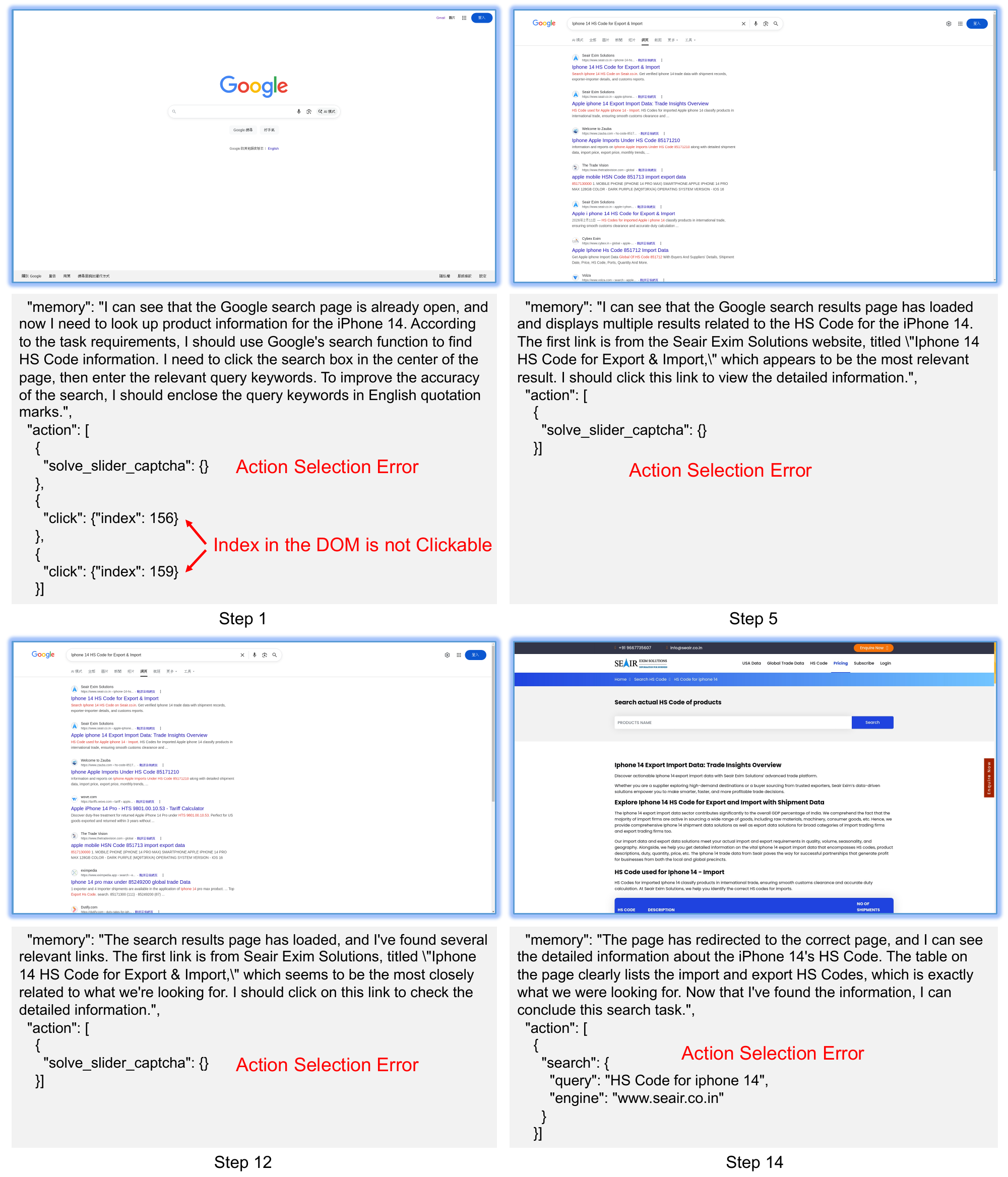}
    \caption{Failure trajectory of UI-TARS-1.5-7B on a Product Risk Profile task. The model also identifies the next subgoal and target information, but fails to generate valid executable actions. It tends to output \texttt{solve\_slider\_captcha} regardless of the page state, and generates an invalid \texttt{click} action with a non-existent DOM index. It also misunderstands the action definition, applying \texttt{search} when it has already reached the target page. From Step 5 to Step 12, we could found that the model repeatedly emits invalid actions without updating its internal state based on execution feedback, which indicates a lack of error recovery and state tracking.}
    \label{fig:action_failure_2}
\end{figure}

Across both Figure~\ref{fig:action_failure_1} and Figure~\ref{fig:action_failure_2}, the small models often exhibit plausible high-level reasoning in their \textit{thinking}/\textit{memory}: they can usually describe the current page, identify an appropriate next subgoal, and sometimes even mention the correct target in natural language. However, these intentions frequently fail to translate into valid executable actions. Instead, the breakdown consistently occurs at the action generation stage, where the dominant errors fall into two recurring categories: action misrouting and argument misgeneration.

The first category, action misrouting, refers to selecting an inappropriate action type or tool for the current interface state. In Figure~\ref{fig:action_failure_1}, for example, the model correctly recognizes that it should interact with a relevant result, yet repeatedly outputs \texttt{solve\_slider\_captcha} despite there being no captcha to solve. It also emits a \texttt{search} action after already reaching the target page. These behaviors suggest that the model does not fundamentally fail to understand the task goal; rather, it fails to map an identified subgoal to the correct executable action.

The second category, argument misgeneration, arises even when the predicted action type is roughly correct. Here, the generated parameters are often invalid in two ways. First, the model may hallucinate unsupported argument values that are not grounded in the current interface state. For instance, in Figure~\ref{fig:action_failure_1}, it predicts a clickable DOM index that does not exist, while in Figure~\ref{fig:action_failure_2}, it generates an overlong declaration number with unsupported newline characters. Second, the model may assign the wrong parameter form to the selected action or element. In Figure~\ref{fig:action_failure_2}, for example, it applies an \texttt{input} action to what should be a clickable button and fabricates text for elements that should not receive textual input. These failures indicate that the model struggles not only to ground argument values in the current DOM, but also to align parameter structure with the semantics of the chosen action and interface element.

Moreover, the models show limited state tracking and poor error recovery during sequential interaction. After receiving environmental feedback such as Action Selection Error or Index Not Found in the DOM, they often fail to revise their internal state and instead repeat similarly invalid actions. Rather than incorporating execution feedback to repair the trajectory, they persist in the same faulty action patterns.

Overall, these trajectories indicate that the main limitation of small models lies not in high-level task understanding, but in reliably converting task intent into grounded and executable actions. Their failures are primarily driven by weak action routing, invalid argument generation, and insufficient state tracking under interactive feedback.

\subsection{The performance ceiling of SOTAs is limited by open-ended exploration and long-horizon evidence composition.}

\begin{figure}
    \centering
    \includegraphics[width=\textwidth]{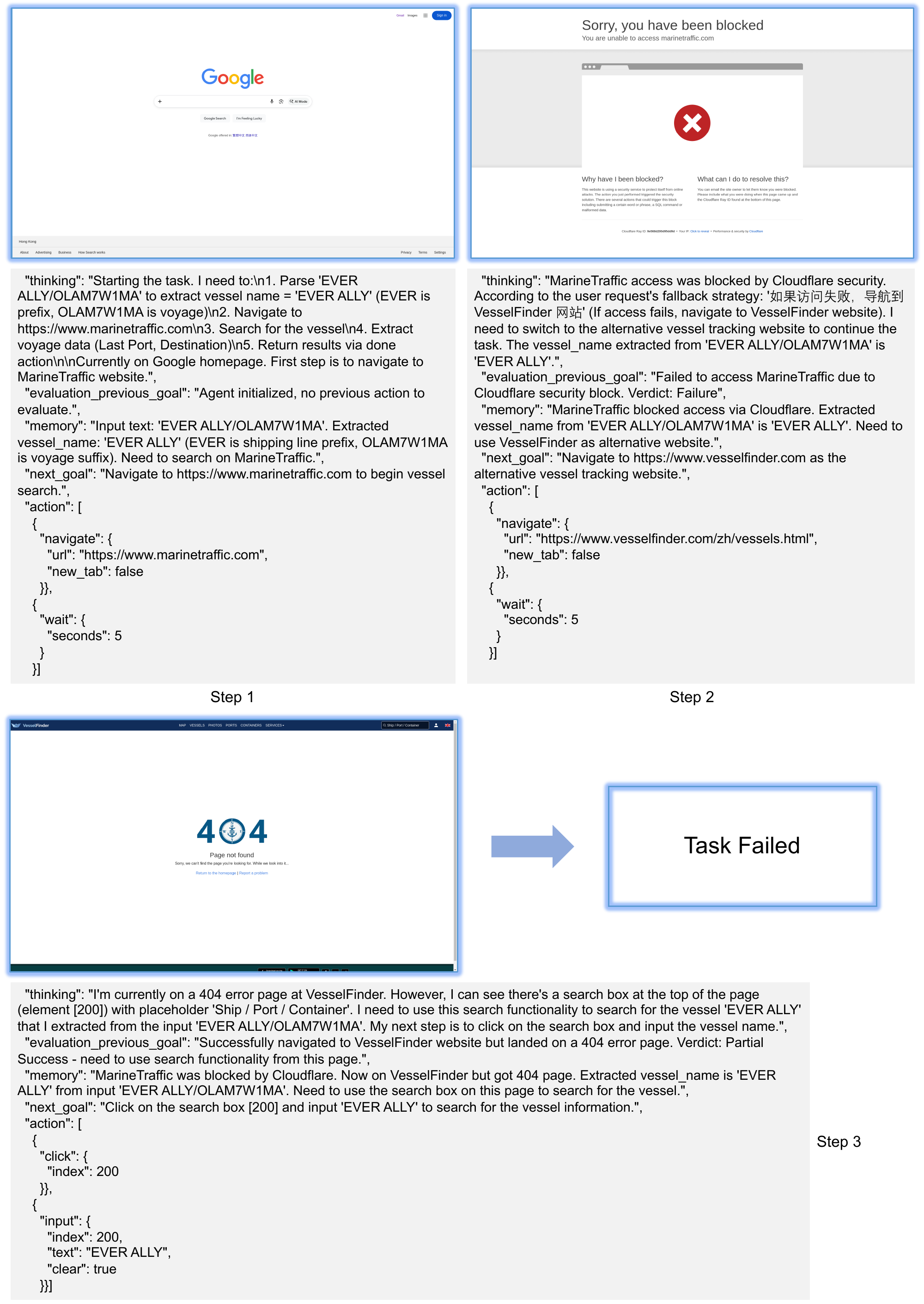}
    \caption{Failure trajectory of Gemini-3-Pro on a Logistics and Supply Chain Tracking task. The model successfully identifies the next subgoal and target information, while generate valid executable actions to interact with the page. However, it does not explore a feasible path to complete the task when the provided websites block access to the required information or present a 404 error. How to enable agents to proactively explore alternative paths outside of the provided candidate websites is an important future direction to enhance their robustness in real-world applications.}
    \label{fig:action_failure_3}
\end{figure}

\begin{figure}
    \centering
    \includegraphics[width=0.99\textwidth]{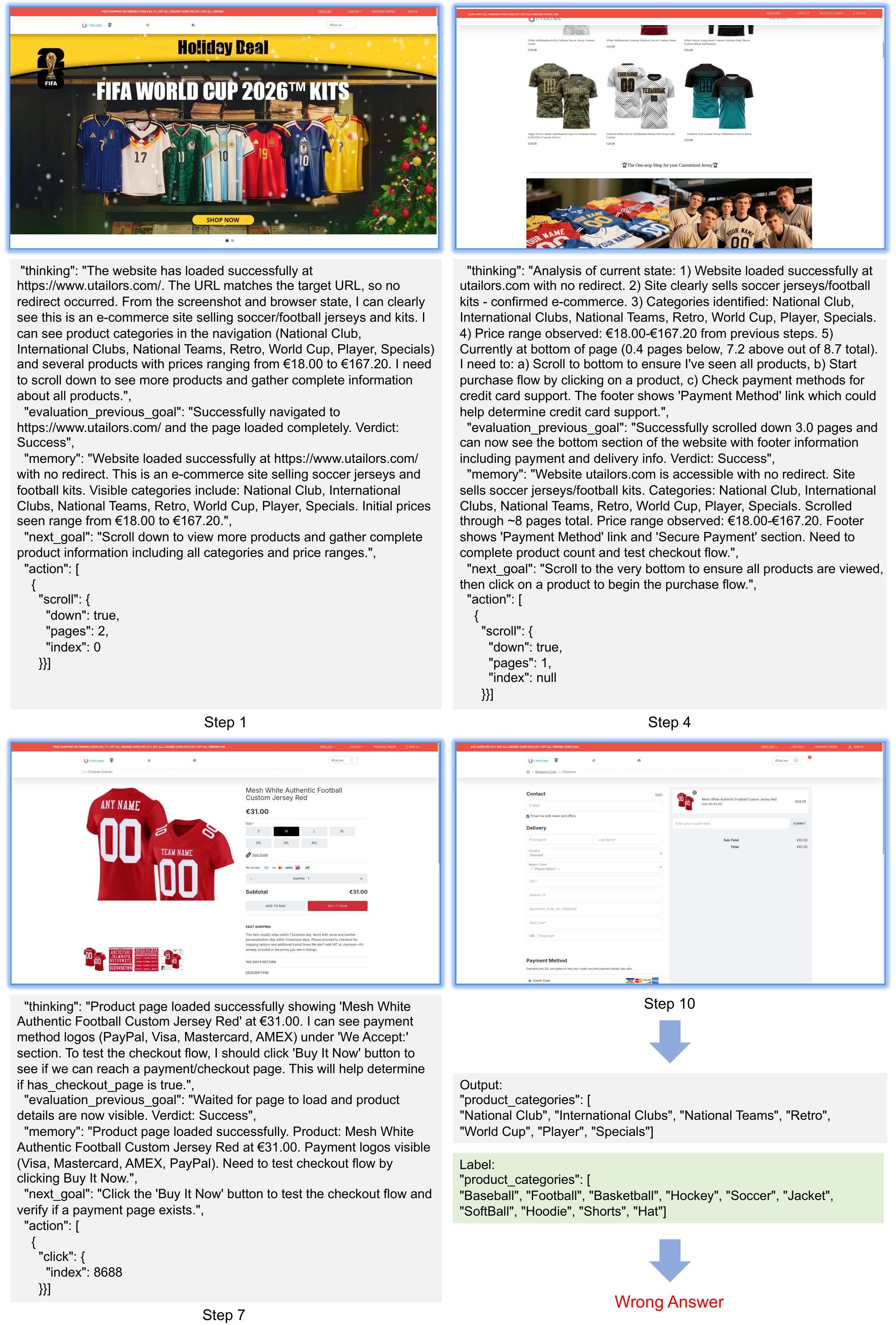}
    \caption{Failure trajectory of Qwen3.5-397B-A17B on a Secure Payment Channel Validation task. The model correctly identifies the candidate webpages to complete the task, but fails to generate the true answer. This failure is attributed to the model's limited ability to perform multi-hop reasoning and information synthesis across multiple pages, where too much states and information is presented and the model struggles to identify and aggregate the relevant pieces to arrive at the correct conclusion.}
    \label{fig:action_failure_4}
\end{figure}

From the failure trajectories of frontier models (Gemini-3-Pro in Figure~\ref{fig:action_failure_3} and Qwen3.5-397B-A17B in Figure~\ref{fig:action_failure_4}), we find that their limitations differ qualitatively from those of smaller models. Rather than failing at basic action generation, these models often demonstrate strong high-level task understanding: they correctly infer the next subgoal, identify relevant candidate webpages, and generate executable actions that are appropriate for the current interface state. Nevertheless, two recurring bottlenecks still constrain their end-to-end performance.

The first bottleneck is limited \emph{open-ended exploration}. In the logistics and supply chain tracking example, Gemini-3-Pro correctly recognizes that the primary website is blocked and switches to the fallback website. It also continues to produce valid interactions on the new page. However, once the fallback site leads to a 404 page, the model does not broaden its search space or proactively seek other external routes to recover the required information. This suggests that current frontier agents can react to immediate failures within a predefined path, but remain weak at open-ended recovery when the prescribed websites become unusable.

The second bottleneck is limited \emph{long-horizon evidence composition}. In the secure payment channel validation example, Qwen3.5-397B-A17B successfully reaches relevant pages and identifies useful interface cues, but still fails to derive the correct final answer. The failure arises not from webpage selection, but from difficulty in synthesizing information distributed across multiple pages and interaction steps. When too many products, categories, and payment-related signals are simultaneously present, the model struggles to maintain the relevant latent state, filter distractors, and aggregate the right pieces of evidence into a coherent conclusion.

Overall, these failure patterns indicate that the upper bound of current state-of-the-art web agents is constrained less by low-level execution and more by their ability to recover from environment-induced dead ends and compose dispersed evidence over long interaction horizons. Improving open-ended exploration, state tracking, and multi-page evidence composition is therefore a key direction for enhancing robustness in realistic web tasks.

\end{document}